\documentclass{article}

\PassOptionsToPackage{numbers, compress}{natbib}
\usepackage[final]{nips_2018}
\usepackage{enumerate}
\usepackage[OT1]{fontenc}
\usepackage{amsmath,amssymb}
\usepackage{wrapfig,lipsum}

\usepackage[usenames]{color}
\usepackage{multirow}
\usepackage[colorlinks,
            linkcolor=red,
            anchorcolor=blue,
            citecolor=blue
            ]{hyperref}

\usepackage{smile}
\usepackage{mathrsfs}

\usepackage[utf8]{inputenc} % allow utf-8 input
\usepackage[T1]{fontenc}    % use 8-bit T1 fonts
\usepackage{hyperref}       % hyperlinks
\usepackage{url}            % simple URL typesetting
\usepackage{booktabs}       % professional-quality tables
\usepackage{amsfonts}       % blackboard math symbols
\usepackage{nicefrac}       % compact symbols for 1/2, etc.
\usepackage{microtype}      % microtypography
\usepackage{smile}
\usepackage{enumitem}

\usepackage[compact]{titlesec}
\usepackage{sidecap}

\title{Stochastic Nested Variance Reduction for \\ Nonconvex Optimization}

\newcommand{\la}{\langle}
\newcommand{\ra}{\rangle}

\def \CC {\textcolor{red}}

\def \xnew {\xb_{p\cdot T_K+j+1}}
\def \xold {\xb_{p\cdot T_K+j}}
\def \hc {\hb_{p\cdot T_K+j}}
\def \Kref {\xb_{p\cdot T_K}}
\def \vkm {\vb_{p\cdot T_K}}
\def \vold {\vb_{p\cdot T_K+j}}

\def \tO {\tilde{O}}
\def \algname {\text{SNVRG}}
\def \df {\nabla f}
\def \dF {\nabla F}

\def \cnew {c^{(K)}_{j+1}}
\def \cold {c^{(K)}_{j}}

\def \ES {\EE}

\def \Ep {\EE}

\def \start {\text{start}}
\def \fin {\text{end}}
\def \con {C}
\def \fon {q}
\def \out {\text{out}}

\def \new {\text{new}}
\def \bbatch {B}
\def \baseb {B}
\def \xbt {\xb^{(k-1)}}
\def \vbt {\vb^{(k-1)}}
\def \xbn {\xb^{(k)}}
\def \vbn {\vb^{(k)}}
\def \cV {\sigma^2}
\def \new {\text{new}}
\def \old {\text{old}}

\author{
  Dongruo Zhou\\
  Department of Computer Science\\
  University of California, Los Angeles\\
  Los Angeles, CA 90095 \\
  \texttt{drzhou@cs.ucla.edu} \\
  \And
  Pan Xu \\
  Department of Computer Science\\
  University of California, Los Angeles\\
  Los Angeles, CA 90095\\
  \texttt{panxu@cs.ucla.edu} \\
  \And
  Quanquan Gu \\
  Department of Computer Science\\
  University of California, Los Angeles\\
  Los Angeles, CA 90095\\
  \texttt{qgu@cs.ucla.edu}
}

\begin{document}

\maketitle

\begin{abstract}
  We study finite-sum nonconvex optimization problems, where the objective function is an average of $n$ nonconvex functions. We propose a new stochastic gradient descent algorithm based on nested variance reduction. Compared with conventional stochastic variance reduced gradient (SVRG) algorithm that uses two reference points to construct a semi-stochastic gradient with diminishing variance in each iteration, our algorithm uses $K+1$ nested reference points to build a semi-stochastic gradient to further reduce its variance in each iteration. For smooth nonconvex functions, the proposed algorithm converges to an $\epsilon$-approximate first-order stationary point (i.e., $\|\nabla F(\mathbf{x})\|_2\leq \epsilon$) within $\tilde O(n\land \epsilon^{-2}+\epsilon^{-3}\land n^{1/2}\epsilon^{-2})$\footnote{$\tilde O(\cdot)$ hides the logarithmic factors, and $a\land b$ means $\min(a,b)$.} number of stochastic gradient evaluations. This improves the best known gradient complexity of SVRG $O(n+n^{2/3}\epsilon^{-2})$ and that of SCSG $O(n\land \epsilon^{-2}+\epsilon^{-10/3}\land n^{2/3}\epsilon^{-2})$. For gradient dominated functions, our algorithm also achieves better gradient complexity than the state-of-the-art algorithms. Thorough experimental results on different nonconvex optimization problems back up our theory.
\end{abstract}

%\CC{how about changing ``deep" to ``nested"?}

\section{Introduction}

We study the following nonconvex finite-sum problem
\begin{align}\label{intro_1}
    \min_{\xb \in \RR^d} F(\xb):=\frac{1}{n}\sum_{i=1}^n f_i(\xb),
\end{align}
where each component function $f_i:\RR^d \rightarrow \RR$ has $L$-Lipschitz continuous gradient but may be nonconvex. A lot of machine learning problems fall into \eqref{intro_1} such as empirical risk minimization (ERM) with nonconvex loss. %We aim to find an $\epsilon$-accurate solution $\xb$ of \eqref{intro_1}, a.k.a., $F(\xb) - \inf_{\yb \in \RR^d}F(\yb) \leq \epsilon$, which is avaliable in convex optimization. However, 
Since finding the global minimum of \eqref{intro_1} is general NP-hard \citep{hillar2013most}, we instead aim at finding an $\epsilon$-approximate stationary point $\xb$, which satisfies $\|\dF(\xb)\|_2 \leq \epsilon$, where $\nabla F(\xb)$ is the gradient of $F(\xb)$ at $\xb$, and $\epsilon>0$ is the accuracy parameter. 

In this work, we mainly focus on first-order algorithms, which only need the function value and gradient evaluations. We use \textit{gradient complexity}, the number of stochastic gradient evaluations, to measure the convergence of different first-order algorithms.\footnote{While we use gradient complexity as in \cite{lei2017non} to present our result, it is basically the same if we use incremental first-order oracle (IFO) complexity used by \cite{Reddi2016Stochastic}. In other words, these are directly comparable.} %The most well-studied first-order methods are \emph{Gradient Descent} (GD) and \emph{Stochastic Gradient Descent} (SGD). %,which are mainly considered in convex problems before. 
For nonconvex optimization, it is well-known that \emph{Gradient Descent} (GD) can converge to an $\epsilon$-approximate stationary point with $O(n\cdot\epsilon^{-2})$ \citep{Nesterov2014Introductory} number of stochastic gradient evaluations. It can be seen that GD needs to calculate the full gradient at each iteration, which is a heavy load when $n\gg 1$. \emph{Stochastic Gradient Descent} (SGD) has $ O(\epsilon^{-4})$ gradient complexity to an $\epsilon$-approximate stationary point under the assumption that the stochastic gradient has a bounded variance \citep{ghadimi2016accelerated}. While SGD only needs to calculate a mini-batch of stochastic gradients in each iteration, due to the noise brought by stochastic gradients, its gradient complexity has a worse dependency on $\epsilon$. %Various \emph{acceleration} techniques \citep{}\CC{XX} have been developed 
In order to improve the dependence of the gradient complexity of SGD on $n$ and $\epsilon$ for nonconvex optimization, %variance reduction technique was proposed. 
%Perhaps the most effective acceleration technique for nonconvex optimization with provable better convergence guarantee is \emph{Variance reduction}.  %is one of these acceleration techniques which 
variance reduction technique was firstly proposed in  \cite{roux2012stochastic,johnson2013accelerating,xiao2014proximal,defazio2014saga, mairal2015incremental, bietti2017stochastic,shalev2013stochastic,defazio2014finito,harikandeh2015stopwasting, nguyen2017sarah, nguyen2017stochastic} for convex finite-sum optimization. Representative algorithms include Stochastic Average Gradient (SAG) \citep{roux2012stochastic}, Stochastic Variance Reduced Gradient (SVRG) \citep{johnson2013accelerating}, SAGA \citep{defazio2014saga}, Stochastic Dual Coordinate Ascent (SDCA) \citep{shalev2013stochastic}, Finito \citep{defazio2014finito}, Batching SVRG \citep{harikandeh2015stopwasting} and SARAH \citep{nguyen2017sarah, nguyen2017stochastic}, to mention a few. The key idea behind variance reduction is that the gradient complexity can be saved if the algorithm use history information as \emph{reference}. 
For instance, one representative variance reduction method is SVRG, which is based on a semi-stochastic gradient that is defined by two reference points. Since the the variance of this semi-stochastic gradient will diminish when the iterate gets closer to the minimizer, it therefore accelerates the convergence of stochastic gradient method. Later on, \citet{harikandeh2015stopwasting} proposed Batching SVRG which also enjoys fast convergence property of SVRG without computing the full gradient.
\iffalse computes the full gradient $\dF(\tilde{\xb})$ every $m$ iterations (called one epoch), and using a semi-stochastic gradient $\df_i(\xb) - \df_i(\tilde{\xb})+\dF(\tilde{\xb})$ instead of full gradient in GD or mini-batch stochastic gradient in SGD, because the the variance of this semi-stochastic gradient will decay when the iterate gets closer to the minimizer. 
It has been shown that SVRG only needs $\tO(n+\kappa)$ stochastic gradient computations to achieve an $\epsilon$-
\fi
The convergence of SVRG under nonconvex setting was first analyzed in \cite{garber2015fast,shalev2016sdca}, where $F$ is still convex but each component function $f_i$ can be nonconvex. The analysis for the general nonconvex function $F$ was done by \cite{Reddi2016Stochastic, allen2016variance}, which shows that SVRG can converge to an $\epsilon$-approximate stationary point with $O(n^{2/3}\cdot \epsilon^{-2})$ number of stochastic gradient evaluations. This result is strictly better than that of GD. Recently, \citet{lei2017non} proposed a \emph{Stochastically Controlled Stochastic Gradient} (SCSG) based on variance reduction, which further reduces the gradient complexity of SVRG to $O(n\land \epsilon^{-2}+\epsilon^{-10/3}\land (n^{2/3}\epsilon^{-2}))$. This result outperforms both GD and SGD strictly.
To the best of our knowledge, this is the state-of-art gradient complexity under the smoothness (i.e., gradient lipschitz) and bounded stochastic gradient variance assumptions. A natural and long standing question is:
\begin{center}
\emph{Is there still room for improvement in nonconvex finite-sum optimization without making additional assumptions beyond smoothness and bounded stochastic gradient variance?}
\end{center}
In this paper, we provide an affirmative answer to the above question, by
showing that the dependence on $n$ in the gradient complexity of SVRG \citep{Reddi2016Stochastic, allen2016variance} and SCSG \citep{lei2017non} can be further reduced. 
We propose a novel algorithm namely \textit{Stochastic Nested Variance-Reduced Gradient descent} (SNVRG). Similar to SVRG and SCSG, our proposed algorithm works in a multi-epoch way. Nevertheless, the technique we developed is highly nontrivial. At the core of our algorithm is the multiple reference points-based variance reduction technique in each iteration. %which gives rise to a semi-stochastic gradient with faster decaying variance. 
In detail, inspired by SVRG and SCSG, which uses two reference points to construct a semi-stochastic gradient with diminishing variance, our algorithm uses $K+1$ reference points to construct a semi-stochastic gradient, whose variance decays faster than that of the semi-stochastic gradient used in SVRG and SCSG. %Both theory and experiments demonstrate the significant advantage of our algorithm against the state-of-the-art algorithms. 
\subsection{Our Contributions} 
Our major contributions are summarized as follows:
\begin{itemize}[leftmargin=*]
    \item We propose a stochastic nested variance reduction technique for stochastic gradient method, which reduces the dependence of the gradient complexity on $n$ compared with SVRG and SCSG. %In fact, GD, SGD , SVRG and SCSG can be seen as can be specialized as GD/SGD/SVRG/SCSG under different parameter settings. 
    \item We show that our proposed algorithm is able to achieve an $\epsilon$-approximate stationary point with $\tO(n\land \epsilon^{-2}+\epsilon^{-3}\land n^{1/2}\epsilon^{-2})$ stochastic gradient evaluations, which outperforms all existing first-order algorithms such as GD, SGD, SVRG and SCSG. 
    \item As a by-product, when $F$ is a $\tau$-gradient dominated function, a variant of our algorithm can achieve an $\epsilon$-approximate global minimizer (i.e., $F(\xb) - \min_\yb F(\yb) \leq \epsilon$) within $\tO\big(n\wedge\tau\epsilon^{-1}+\tau (n\wedge \tau\epsilon^{-1})^{1/2}\big)$ stochastic gradient evaluations, which also outperforms the state-of-the-art. 
\end{itemize}

%The remainder of this paper is organized as follows: 

\subsection{Additional Related Work}
Since it is hardly possible to review the huge body of literature on convex and nonconvex optimization due to space limit, here we review some additional most related work on accelerating nonconvex (finite-sum) optimization. 

\begin{wrapfigure}{r}{5.0cm}
	%\centering
    \includegraphics[width=0.36\textwidth]{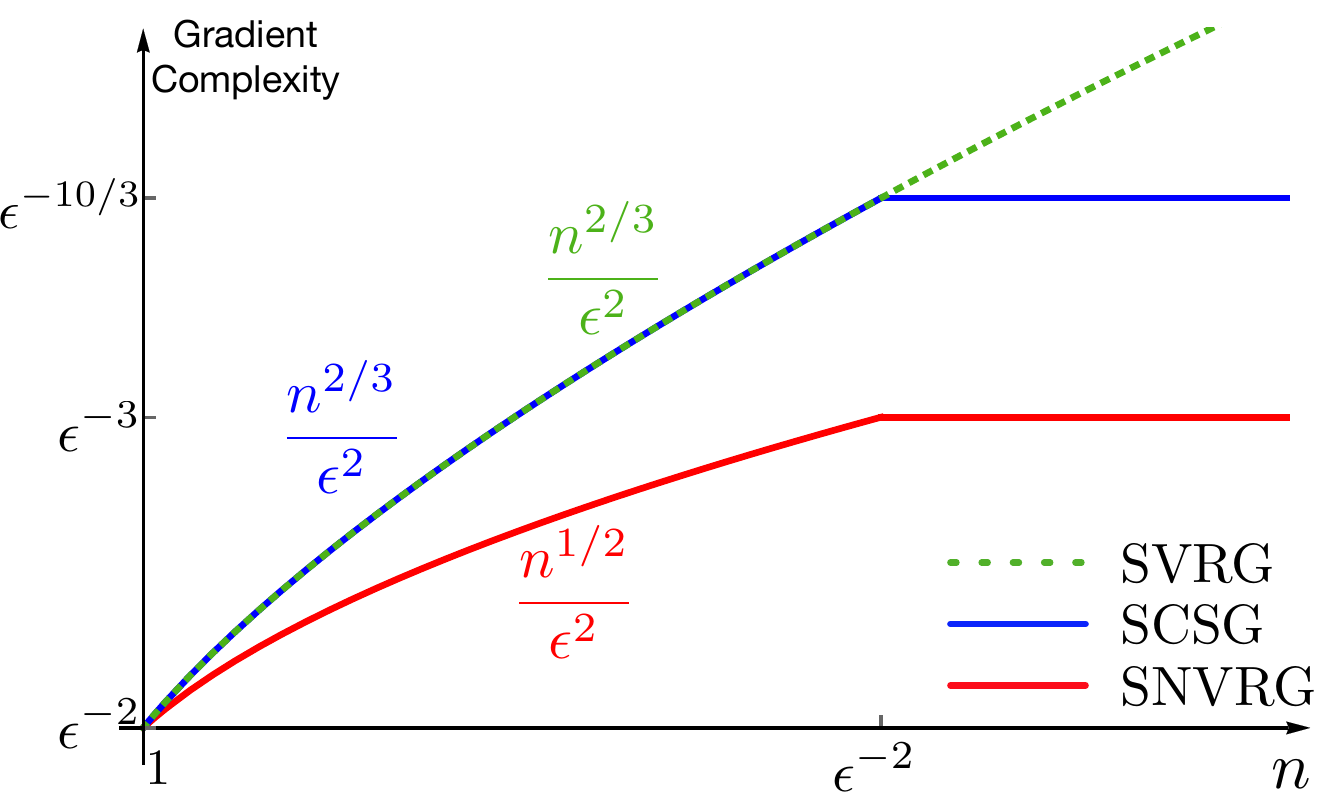}
	\caption{Comparison of gradient complexities.}
	\label{fig:comp_gradient_comple}
	%\vskip -0.1in
\end{wrapfigure}
\noindent\textbf{Acceleration by high-order smoothness assumption}
With only Lipschitz continuous gradient assumption, \citet{carmonlower} showed that the lower bound for both deterministic and stochastic algorithms to achieve an $\epsilon$-approximate stationary point is $\Omega(\epsilon^{-2})$. With high-order smoothness assumptions, i.e., Hessian Lipschitzness, Hessian smoothness etc., a series of work have shown the existence of acceleration. For instance, \citet{Agarwal2017Finding} gave an algorithm based on Fast-PCA which can achieve 
an $\epsilon$-approximate stationary point with gradient complexity $\tO(n\epsilon^{-3/2}+n^{3/4}\epsilon^{-7/4})$  \citet{Carmon2016Accelerated, carmon2017convex} showed two algorithms based on finding exact or inexact negative curvature which can achieve 
an $\epsilon$-approximate stationary point with gradient complexity $\tO(n\epsilon^{-7/4})$. In this work, we only consider gradient Lipschitz without assuming Hessian Lipschitz or Hessian smooth. Therefore, our result is not directly comparable to the methods in this category.

\noindent\textbf{Acceleration by momentum}
The fact that using momentum is able to accelerate algorithms has been shown both in theory and practice in convex optimization \citep{polyak1964some,nesterov2005smooth,hu2009accelerated,lan2012optimal,ghadimi2012optimal,Nesterov2014Introductory,lin2015universal,allen2017katyusha}. However, there is no evidence that such acceleration exists in nonconvex optimization with only Lipschitz continuous gradient assumption \citep{ghadimi2016accelerated,li2015accelerated, paquette2017catalyst,li2017convergence,lan2017optimal}. If $F$ satisfies $\lambda$-strongly nonconvex, i.e., $\nabla^2 F \succeq -\lambda \Ib$, \citet{allen2017natasha} proved that Natasha $1$, an algorithm based on nonconvex momentum, is able to find an $\epsilon$-approximate stationary point in $\tO(n^{2/3}L^{2/3}\lambda^{1/3}\epsilon^{-2})$. Later, \citet{allen2017natasha} further showed that Natasha 2, an online version of Natasha 1, is able to achieve an $\epsilon$-approximate stationary point within $\tO(\epsilon^{-3.25})$ stochastic gradient evaluations\footnote{In fact, Natasha 2 is guaranteed to converge to an $(\epsilon,\sqrt{\epsilon})$-approximate second-order stationary point with $\tilde O(\epsilon^{-3.25})$ gradient complexity, which implies the convergence to an $\epsilon$-approximate stationary point.}. 
%\CC{may need to cite guanghui lan, zhouchen lin's NIPS2016, yingbin liang's last ICML, nonconvex catelyst paper, and so on here, to avoid enemy}

After our paper was submitted to NIPS and released on arXiv, a paper \cite{fang2018spider} was released on arXiv after our work, which independently proposes a different algorithm and achieves the same convergence rate for finding an $\epsilon$-approximate stationary point.

To give a thorough comparison of our proposed algorithm with existing first-order algorithms for nonconvex finite-sum optimization, we summarize the gradient complexity of the most relevant algorithms in Table \ref{table:complexity}. We also plot the gradient complexities of different algorithms in Figure \ref{fig:comp_gradient_comple} for nonconvex smooth functions. Note that GD and SGD are always worse than SVRG and SCSG according to Table \ref{table:complexity}. In addition, GNC-AGD and Natasha2 needs additional Hessian Lipschitz condition. Therefore, we only plot the gradient complexity of SVRG, SCSG and our proposed SNVRG in Figure \ref{fig:comp_gradient_comple}.

\begin{table}[t]
\caption{Comparisons on gradient complexity of different algorithms. The second column shows the gradient complexity for a nonconvex and smooth function to achieve an $\epsilon$-approximate stationary point (i.e., $\|\nabla F(\xb)\|_2\leq\epsilon$). The third column presents the gradient complexity for a gradient dominant function to achieve an $\epsilon$-approximate global minimizer (i.e., $F(\xb) - \min_\xb F(\xb) \leq \epsilon$). The last column presents the space complexity of all algorithms.}
\label{table:complexity}
\begin{small}
\begin{center}
\begin{tabular}{cccc}
\toprule
Algorithm &nonconvex & gradient dominant & Hessian Lipschitz\\
%&&&\\
\midrule
GD& $O\big(\frac{n}{\epsilon^{2}}\big)$ & $\tO(\tau n)$& No\\
SGD&$O\big(\frac{1}{\epsilon^{4}}\big)$ & $O\big(\frac{1}{\epsilon^{4}}\big)$ & No\\
SVRG \citep{Reddi2016Stochastic}&$O\big( \frac{n^{2/3}}{  \epsilon^{2}}\big)$&$\tO(n+\tau n^{2/3})$& No\\
SCSG \small{\citep{lei2017non}} & $O\big(\frac{1}{\epsilon^{10/3}}\land \frac{n^{2/3}}{ \epsilon^{2}}\big)$& $\tO\Big(n\wedge\frac{\tau}{\epsilon}+\tau \big(n\wedge \frac{\tau}{\epsilon}\big)^{2/3}\Big)$ & No\\
GNC-AGD \citep{carmon2017convex}&$\tO\big(\frac{n}{\epsilon^{1.75}}\big)$& N/A & Needed\\
%\midrule
Natasha 2  \small{\citep{allen2017natasha}} & $\tO\big(\frac{1}{\epsilon^{3.25}}\big)$  & N/A  &Needed\\
%\midrule 
$\algname$  (this paper)  &$\tO\big(\frac{1}{\epsilon^{3}}\land \frac{n^{1/2}}{\epsilon^{2}}\big)$ & $\tO\Big(n\wedge\frac{\tau}{\epsilon}+\tau \big(n\wedge \frac{\tau}{\epsilon}\big)^{1/2}\Big)$ & No \\
\bottomrule
\end{tabular}
\end{center}
\end{small}
\end{table}

\noindent\textbf{Notation:}  Let $\Ab = [A_{ij}] \in \RR^{d\times d}$ be a matrix and $\xb = (x_1,...,x_d)^{\top} \in \RR^{d}$ be a vector. $\Ib$ denotes an identity matrix. We use $\|\vb\|_2$ to denote the 2-norm of vector $\vb \in \RR^d$. We use $\la\cdot,\cdot\ra$ to represent the inner product of two vectors. %\CC{Define Big O and Big tilde O} 
Given two sequences $\{a_n\}$ and $\{b_n\}$, we write $a_n = O(b_n)$ if there exists a constant $0 < C < +\infty$ such that $a_n \leq C\, b_n$. We write $a_n = \Omega(b_n)$ if there exists a constant $0<C<+\infty$, such that $a_n \geq C\, b_n$. We use notation $\tilde{O}(\cdot)$ to hide logarithmic factors. We also make use of the notation $f_n\lesssim g_n$ ($f_n\gtrsim g_n$) if $f_n$ is less than (larger than) $g_n$ up to a constant. 
We use productive symbol $\prod_{i=a}^b c_i$ to denote $c_ac_{a+1}\dots c_b$. Moreover, if $a>b$, we take the product as 1. We use $\lfloor\cdot \rfloor$ as the floor function. We use $\log(x)$ to represent the logarithm of $x$ to base 2. $a\land b$ is a shorthand notation for $\min(a,b)$.

\section{Preliminaries}
%\CC{definition}

In this section, we present some definitions that will be used throughout our analysis.
\begin{definition}\label{smooth}
A function $f$ is $L$-smooth, if for any $\xb, \yb \in \RR^d$, we have
\begin{align}
    \|\df(\xb) - \df(\yb)\|_2 \leq L\|\xb - \yb\|_2.
\end{align}
\end{definition}
Definition \ref{smooth} implies that if $f$ is $L$-smooth,   we have for any $\xb,\hb\in\RR^d$
\begin{align}
    f(\xb+\hb) \leq f(\xb)+\la \nabla f(\xb), \hb\ra+\frac{L}{2}\|\hb\|_2^2.
\end{align}

%\CC{change the name of the following statement bounded variance}
\begin{definition}\label{def:strong_convex}
A function $f$ is $\lambda$-strongly convex, if for any $\xb, \yb \in \RR^d$, we have
\begin{align}
    f(\xb+\hb) \geq f(\xb)+\la \nabla f(\xb), \hb\ra+\frac{\lambda}{2}\|\hb\|_2^2.
\end{align}
\end{definition}

%\CC{change $\cV$ to $\sigma^2$ to distinguish with Lihua Lei's notation}
%\CC{change the statement of the following assumption}
\begin{definition}\label{boundgra}
A function $F$ with finite-sum structure in \eqref{intro_1} is said to have stochastic gradients with bounded variance $\cV$, if for any $\xb \in \RR^d$, we have
\begin{align}
    \EE_i\|\df_i(\xb) - \dF(\xb)\|_2^2 \leq \cV,
\end{align}
where $i$ a random index uniformly chosen from $[n]$ and $\EE_i$ denotes the expectation over such $i$. 
\end{definition}
$\sigma^2$ is called the upper bound on the variance of stochastic gradients \citep{lei2017non}.
%\CC{add}
\begin{definition}\label{avsmooth}
A function $F$ with finite-sum structure in \eqref{intro_1} is said to have averaged $L$-Lipschitz gradient, if for any $\xb, \yb \in \RR^d$, we have
\begin{align}
    \EE_i\|\nabla f_i(\xb) - \nabla f_i(\yb)\|_2^2 \leq L^2\|\xb - \yb\|_2^2,
\end{align}
where $i$ is a random index uniformly chosen from $[n]$ and $\EE_i$ denotes the expectation over the choice.
\end{definition}
\begin{definition}
We say a function $f$ is lower-bounded by $f^*$ if for any $\xb \in \RR^d$, $f(\xb) \geq f^*$. 
\end{definition}
%\CC{$f^*$ or $F^*$?}

We also consider a class of functions namely gradient dominated functions \citep{polyak1963gradient}, which is formally defined as follows:
\begin{definition}\label{def:gradientdo}
We say function $f$ is $\tau$-gradient dominated if for any $\xb \in \RR^d$, we have
\begin{align}
    f(\xb) - f(\xb^*) \leq \tau\cdot \|\nabla f(\xb)\|_2^2, 
\end{align}
where $\xb^*\in \RR^d$ is the global minimum of $f$.
\end{definition}
Note that gradient dominated condition is  also known as the Polyak-Lojasiewicz (P-L) condition \citep{polyak1963gradient}, and is not necessarily convex. It is weaker than strong convexity as well as other popular conditions that appear in the optimization literature \citep{karimi2016linear}. 

%\CC{add definition of $\lambda$-strongly convex}
\section{The Proposed Algorithm}\label{proposed_algorithm}
In this section, we present our nested stochastic variance reduction algorithm, namely, $\algname$.

\begin{algorithm}[!htbp]
\caption{One-epoch-$\algname$($\xb_0$, $F, K, M, \{T_l\}, \{B_l\}, B$)}\label{algorithm:1}
\begin{algorithmic}[1]
  \STATE \textbf{Input:}  initial point $\xb_0$, function $F$, loop number $K$, step size parameter $M$,  loop parameters $T_l, l \in [K]$, batch parameters $B_l, l \in [K]$, base batch size $B>0$. 
  
  \STATE $\xb^{(l)}_0 \gets \xb_0$, $\gb^{(l)}_0 \gets 0$, $0\leq l \leq K$
  \STATE Uniformly generate index set $I \subset [n]$ without replacement, $|I| = B$
  \STATE $\gb^{(0)}_0 \gets 1/B\sum_{i \in I}\nabla f_{i}(\xb_0)$\label{alg_line:update_g0}
  \STATE $\vb_0  \gets  \sum_{l=0}^{K} \gb_0^{(l)}$
  \STATE $\xb_1 = \xb_0-1/(10M)\cdot \vb_0$
  
  \FOR{$t = 1,...,\prod_{l=1}^K T_l - 1$}

  \STATE $r = \min \{j: 0 = ( t \mod \prod_{l=j+1}^K T_l) ,\ 0\leq j \leq K\}$
  
  \STATE $ \{\xb^{(l)}_t\} \gets \text{Update\_reference\_points}(\{\xb^{(l)}_{t-1}\}, \xb_t,r),0 \leq l \leq K$. 
  \STATE $\{\gb_t^{(l)}\} \gets \text{Update\_reference\_gradients}(\{\gb^{(l)}_{t-1}\}, \{\xb^{(l)}_{t}\}, r),0 \leq l \leq K$.
  
  \STATE$\vb_t  \gets  \sum_{l=0}^{K} \gb_t^{(l)}$
  \STATE $\xb_{t+1} \gets \xb_{t}- 1/(10M)\cdot \vb_t$\label{alg_line:update_gd}
  \ENDFOR
  \STATE $\xb_{\out}\gets$ uniformly random choice from $\{\xb_{t}\}$, where $ 0\leq t <  \prod_{l=1}^K T_l$
  %\STATE $\xb_{\eend} \gets \xb_{\prod_{l=1}^K T_l}$
  \STATE $T=\prod_{l=1}^K T_l$
  \STATE \textbf{Output:} $[\xb_{\out}, \xb_{T}]$

  \hrulefill	

  \STATE \textbf{Function:} Update\_reference\_points($\{\xb_{\old}^{(l)}\},  \xb,r$)
  
  %\STATE $r \gets \min \{j: t = \lfloor t/\prod_{i=j+1}^K T_i\rfloor \cdot \prod_{i=j+1}^K T_i , 0\leq j \leq K\}$
\STATE $\xb_{\new}^{(l)}  \gets \xb_{\old}^{(l)}, 0 \leq l \leq r-1$; $ \xb_{\new}^{(l)}  \gets \xb, r\leq l \leq K$\label{alg_line:update_x}
\RETURN $\{\xb_{\new}^{(l)}\}$

\hrulefill

\STATE \textbf{Function:} Update\_reference\_gradients($\{\gb^{(l)}_{\old}\}, \{\xb^{(l)}_{\new}\}, r$)
\STATE $\gb^{(l)}_{\new} \gets \gb^{(l)}_{\old}, 0 \leq l < r$\label{alg_line:update_grad_small}
\FOR{$r\leq l \leq K$}
  \STATE Uniformly generate index set $I \subset [n]$ without replacement, $|I| = B_l$
  \STATE $\gb_{\new}^{(l)} \gets 1/B_l\sum_{i \in I}\big[\nabla f_{i}(\xb^{(l)}_{\new}) - \nabla f_{i}(\xb^{(l-1)}_{\new})\big]$ \label{alg_line:update_grad}
  \ENDFOR \label{alg_line:update_grad_large}
  \RETURN $ \{\gb_{\new}^{(l)}\}$.
\end{algorithmic}
\end{algorithm}
\noindent\textbf{One-epoch-\algname:}
We first present the key component of our main algorithm, One-epoch-$\algname$, which is displayed in Algorithm \ref{algorithm:1}. The most innovative part of Algorithm \ref{algorithm:1} attributes to the $K+1$ \emph{reference points} and $K+1$ \emph{reference gradients}. Note that when $K=1$, Algorithm \ref{algorithm:1} reduces to one epoch of SVRG algorithm \citep{johnson2013accelerating,Reddi2016Stochastic,allen2016variance}. To better understand our One-epoch $\algname$ algorithm, it would be helpful to revisit the original SVRG which is a special case of our algorithm. For the finite-sum optimization problem in \eqref{intro_1}, the original SVRG takes the following updating formula  
\begin{align*}
    \xb_{t+1} = \xb_t -\eta\vb_t=\xb_t-\eta\big(\nabla F(\tilde\xb)+\nabla f_{i_t}(\xb_t)-\nabla f_{i_t}(\tilde\xb)\big),
\end{align*}
where $\eta>0$ is the step size, $i_t$ is a random index uniformly chosen from $[n]$ and $\tilde\xb$ is a snapshot for $\xb_t$ after every $T_1$ iterations. There are two reference points in the update formula at $\xb_t$: $\xb_t^{(0)}=\tilde\xb$ and $\xb_t^{(1)}=\xb_t$. Note that $\tilde\xb$ is updated every $T_1$ iterations, namely, $\tilde\xb$ is set to be $\xb_t$ only when $(t \mod T_1) = 0$. Moreover, in the semi-stochastic gradient $\vb_t$, there are also two reference gradients and we denote them by $\gb_t^{(0)}=\nabla F(\tilde\xb)$ and $\gb_t^{(1)}=\nabla f_{i_t}(\xb_t)-\nabla f_{i_t}(\tilde\xb)=\nabla f_{i_t}(\xb_t^{(1)})-\nabla f_{i_t}(\xb_t^{(0)})$. %The references points and gradients used in SVRG are illustrated in Figure \ref{fig:illu_svrg} with $K=1$.

\begin{wrapfigure}{r}{4.8cm}
    \vskip-0.1in
	\centering
	\includegraphics[width=1.05\linewidth]{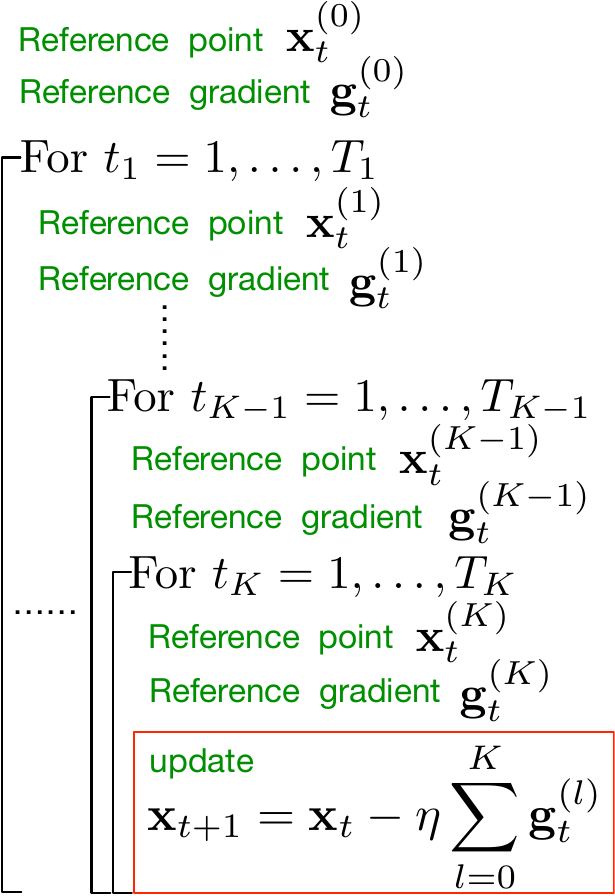}
	\caption{Illustration of reference points and gradients.\label{fig:illu_snvrg}}
	\vskip -0.15in
\end{wrapfigure}

Back to our One-epoch-$\algname$, we can define similar reference points and reference gradients as that in the special case of SVRG. Specifically, for $t=0,\ldots, \prod_{l=1}^K T_l-1$, each point $\xb_t$ has $K+1$ reference points $\{\xb_t^{(l)}\}, l=0,\ldots,K$, which is set to be $\xb_t^{(l)} = \xb_{t^l}$ with index $t^l$ defined as
\begin{align}\label{referencepoints}
t^l = \bigg\lfloor\frac{t}{\prod_{k=l+1}^K T_k}\bigg\rfloor\cdot\prod_{k=l+1}^K T_k.
\end{align}
Specially, note that we have $\xb_t^{(0)}=\xb_0$ and $\xb_t^{(K)}=\xb_t$ for all $t=0,\ldots, \prod_{l=1}^K T_l-1$. Similarly, $\xb_t$ also has $K+1$ reference gradients $\{\gb_t^{(l)}\}$, which can be defined based on the reference points $\{\xb_t^{(l)}\}$:
\begin{align}\label{eq:update_gt}
\begin{split}
    &\gb_t^{(0)} = \frac{1}{B}\sum_{i \in I}\nabla f_{i}(\xb_0),\qquad \gb_t^{(l)} = \frac{1}{B_l}\sum_{i \in {I_l}}\big[\nabla f_{i}(\xb_t^{(l)}) - \nabla f_{i}(\xb_t^{(l-1)})\big], l=1,\ldots,K,
\end{split}
\end{align}
where $I, I_l$ are random index sets with $|I| = B, |I_l| = B_l$ and are uniformly generated from $[n]$ without replacement. Based on the reference points and reference gradients, we then update $\xb_{t+1} = \xb_t - 1/(10M)\cdot \vb_t$, where $\vb_t = \sum_{l=0}^K \gb_t^{(l)}$ and $M$ is the step size parameter. The illustration of reference points and gradients of $\algname$ is displayed in Figure \ref{fig:illu_snvrg}. %\CC{Figure 1 (a)}

We remark that it would be a huge waste for us to re-evaluate $\gb_t^{(l)}$ at each iteration. Fortunately, due to the fact that each reference point is only updated after a long period, we can maintain $\gb_t^{(l)} = \gb_{t-1}^{(l)}$ and only need to update $\gb_t^{(l)}$ when $\xb_t^{(l)}$ has been updated as is suggested by Line~\ref{alg_line:update_grad} in Algorithm \ref{algorithm:1}. %This trick will save a lot of gradient computation and we formally have the following proposition:
% \begin{proposition}
% Given $w,s$ which satisfy $0 \leq w\cdot \prod_{j=s+1}^K T_j < (w+1)\cdot \prod_{j=s+1}^K T_j < T$, then for any $t,t'$ which satisfy  $w\cdot \prod_{j=s+1}^K T_j \leq t<t' < (w+1)\cdot \prod_{j=s+1}^K T_j$, we have $\xb_{t}^{(s)} =\xb_{t'}^{(s)}= \xb_{w\cdot \prod_{j=s+1}^K T_j}$. Moreover, for any $i \leq s$, we have $\gb_{t}^{(i)} = \gb_{t'}^{(i)}$.
% \end{proposition}

% \begin{wrapfigure}{r}{8.40cm}
% %\begin{figure}[t]
%     %\vskip -0.2in 
% 	\centering
% 	\subfigure[SVRG]{\includegraphics[width=0.4\linewidth]{arXiv/figures/svrg}
% 	\label{fig:illu_svrg}}
% 	\subfigure[SNVRG]{
% 	\includegraphics[width=0.4\linewidth]{arXiv/figures/SNVRG}
% 	\label{fig:illu_snvrg}}
% 	\caption{Illustration of reference points and gradients in SVRG and SNVRG.}
% 	\label{fig:illustration}
% 	%\vskip -0.5in 
% %\end{figure}
% \end{wrapfigure}
\noindent\textbf{SNVRG:}
Using One-epoch-$\algname$ (Algorithm \ref{algorithm:1}) as a building block, we now present our main algorithm: Algorithm \ref{algorithm:general} for nonconvex finite-sum optimization to find an $\epsilon$-approximate stationary point. %and %Algorithm \ref{algorithm:gradientdom} for gradient dominated functions to find an $\epsilon$-accurate solution. 
%In Algorithm \ref{algorithm:general}, we treat One-epoch-$\algname$ as a black box. 
At each iteration of Algorithm \ref{algorithm:general}, it executes One-epoch-$\algname$ (Algorithm \ref{algorithm:1}) which takes $\zb_{s-1}$ as its input and  outputs $[\yb_s, \zb_s]$. We choose $\yb_{\text{out}}$ as the output of Algorithm \ref{algorithm:general} uniformly from $\{\yb_s\}$, for $s=1,\ldots,S$.

%Based on the One-epoch-$\algname$ Algorithm, we present our main algorithm, Stochastic Nested Variance Reduced Gradient ($\algname$), in Algorithm \ref{algorithm:general}. 

\noindent\textbf{SNVRG-PL:} In addition, when function $F$ in \eqref{intro_1} is gradient dominated as defined in Definition \ref{def:gradientdo} (P-L condition), it has been proved that the global minimum can be found by SGD \citep{karimi2016linear}, SVRG \citep{Reddi2016Stochastic} and SCSG \citep{lei2017non} very efficiently. Following a similar trick used in \cite{Reddi2016Stochastic}, we present Algorithm \ref{algorithm:gradientdom} on top of Algorithm \ref{algorithm:general}, to find the global minimum in this setting. We call Algorithm \ref{algorithm:gradientdom} SNVRG-PL, because gradient dominated condition is also known as Polyak-Lojasiewicz (PL) condition \citep{polyak1963gradient}. %The modified algorithm is shown in , which follows .

\noindent\textbf{Space complexity}: We briefly compare the space complexity between our algorithms and other variance reduction based algorithms. SVRG and SCSG needs $O(d)$ space complexity to store one reference gradient, SAGA \citep{defazio2014saga} needs to store reference gradients for each component functions, and its space complexity is $O(nd)$ without using any trick. For our algorithm SNVRG, we need to store $K$ reference gradients, thus its space complexity is $O(Kd)$. In our theory, we will show that $K=O(\log \log n)$. Therefore, the space complexity of our algorithm is actually $\tilde O(d)$, which is almost comparable to that of SVRG and SCSG.

%%%%%%%%%%%%%%%%%%%%%%%seperate%%%%%%%%%%%%
% \begin{algorithm*}[ht]
% \caption{SNVRG($\zb_0, F, K, M, \{T_l\}, \{B_l\},  B, S$)}\label{algorithm:general}
% \begin{algorithmic}[1]
%   \STATE \textbf{Input:} initial point $\zb_0$, function $F$, $K$, $M$, $\{T_l\}$, $\{B_l\}$, batch $B$, $S$.
%   \FOR{$s=1,\dots,S$}
%   \STATE $[\yb_s, \zb_s] \gets \text{One-epoch-}\algname(\zb_{s-1},F, K, M, \{T_l\}, \{B_l\}, B)$ \hfill{$\triangleright$ Algorithm \ref{algorithm:1}}
%   \ENDFOR
%   \STATE \textbf{Output:} Uniformly choose $\yb_{\text{out}}$ from $\{\yb_s\},  1\leq s \leq S$.
%  \end{algorithmic}
% \end{algorithm*}
% \begin{algorithm*}[ht]
% \caption{SNVRG-PL($\zb_0,F, K, M, \{T_l\}, \{B_l\},  B, S, U$)}\label{algorithm:gradientdom}
% \begin{algorithmic}[1]
%   \STATE \textbf{Input:}initial point $\zb_0$, function $F$, $K$, $M$, $\{T_l\}$, $\{B_l\}$, batch $B$, $S$, $U$.
%   \FOR{$u=1,\dots,U$}
%   \STATE $\zb_u = \algname(\zb_{u-1},F, K, M, \{T_l\}, \{B_l\},  B, S)$ 
%   \hfill{$\triangleright$ Algorithm \ref{algorithm:general}}
%   \ENDFOR
%   \STATE \textbf{Output:} $\zb_{\text{out}} = \zb_U$.
%  \end{algorithmic}
% \end{algorithm*}

%%%%%%%%%%%%%%%%%%%%%%%minipage%%%%%%%%%%%%
\begin{minipage}[t]{0.49\linewidth}
\begin{algorithm}[H]
\caption{SNVRG}
\label{algorithm:general}
\begin{algorithmic}[1]
  \STATE \textbf{Input:} initial point $\zb_0$, function $F$, $K$, $M$, $\{T_l\}$, $\{B_l\}$, batch $B$, $S$.
  \FOR{$s=1,\dots,S$}\STATE denote $\cP=(F,K, M, \{T_l\}, \{B_l\},B)$
  \STATE $[\yb_s, \zb_s]$ $\gets$$\text{One-epoch-}\algname(\zb_{s-1}, \cP)$% where $\cP=(F,K, M, \{T_l\}, \{B_l\},B)$ is the tuple of all the rest parameters.
  \ENDFOR
  \STATE \textbf{Output:} Uniformly choose $\yb_{\text{out}}$ from $\{\yb_s\}$.
 \end{algorithmic}
\end{algorithm}
\end{minipage}
\hfil
\begin{minipage}[t]{0.49\linewidth}
\begin{algorithm}[H]
\caption{SNVRG-PL}
\label{algorithm:gradientdom}
\begin{algorithmic}[1]
  \STATE \textbf{Input:} initial point $\zb_0$, function $F$, $K$, $M$, $\{T_l\}$, $\{B_l\}$, batch $B$, $S$, $U$.
%  \STATE $\bbatch_s = \bbatch, 1 \leq s \leq S$
  \FOR{$u=1,\dots,U$}
  \STATE denote $\cQ=(F,K, M, \{T_l\}, \{B_l\}, B, S)$
  \STATE $\zb_u = \algname(\zb_{u-1},\cQ)$ 
  \ENDFOR
  \STATE \textbf{Output:} $\zb_{\text{out}} = \zb_U$.
 \end{algorithmic}
\end{algorithm}
\end{minipage}

\section{Main Theory}\label{sec:theory}
In this section, we provide the convergence analysis of $\algname$. %At the core of the convergence proof is the analysis of One-epoch-$\algname$ (Algorithm \ref{algorithm:1}), which is illustrated in the following subsection. 

%\subsection{Convergence Analysis of One-epoch-\algname}
\subsection{Convergence of SNVRG}

We first analyze One-epoch-$\algname$ (Algorithm \ref{algorithm:1}) and provide a particular choice of parameters.

\begin{lemma}\label{choosek_single}
Suppose that $F$ has averaged $L$-Lipschitz gradient, in Algorithm \ref{algorithm:1}, suppose $B \geq 2$ and let the number of nested loops be $K  =  \log \log B$. Choose the step size parameter as $M = 6L$. For the loop and batch parameters, let $T_1=2, B_1=6^K\cdot B$ and 
\begin{align}
    T_l = 2^{2^{l-2}},\qquad B_l  = 6^{K-l+1}\cdot B/2^{2^{l-1}},\notag
\end{align}
for all $2 \leq l \leq K$. Then the output of Algorithm \ref{algorithm:1} $[\xb_{\text{out}}, \xb_{T}]$ satisfies
\begin{align}
    &\ES \|\dF(\xb_{\text{out}})\|_2^2\leq \con\bigg(\frac{L}{\baseb^{1/2}}\cdot\EE\big[F(\xb_{0})- F(\xb_{T})\big]+ \frac{\cV}{\baseb}\cdot \ind (\baseb < n) \bigg)\label{choosek_single_1}
\end{align}
within $1\lor (7\baseb \log^3 \baseb)$ stochastic gradient computations, where $T=\prod_{l=1}^K T_l$, $\con = 600$ is a constant and $\ind(\cdot)$ is the indicator function. 
%Note that the choice of $K, M, \{T_i\}, \{B_i\}$ is only related to $B$ and $L$. Therefore, we denote the above choice of parameters as $(K, M, \{T_i\}, \{B_i\}) = Opt(B, L)$.
\end{lemma}

\iffalse
\begin{remark}
An very interesting observation about the parameter choice in One-epoch-$\algname$ is that it separates the parameters into two groups. One group containing only the step size parameter $M$ is only dependent on the Lipschitz constant $L$; another group consisting of loop parameters $\{T_i\}$, batch parameters $\{B_i\}$ and the number of nested loops $K$ is only related to the base batch size $B$. 
\end{remark}
\fi

The following theorem shows the gradient complexity for Algorithm \ref{algorithm:general} to find an $\epsilon$-approximate stationary point with a constant base batch size $\bbatch$.
 
\begin{theorem}\label{maintheorem}
Suppose that $F$ has averaged $L$-Lipschitz gradient and stochastic gradients with bounded variance $\sigma^2$. In Algorithm \ref{algorithm:general}, let $\bbatch = n \land (2\con\cV/\epsilon^2)$ , $S=1 \lor (2\con L \Delta_F/(\bbatch^{1/2}\epsilon^2))$ and $\con= 600$. The rest parameters $(K,M,\{B_l\},\{T_l\})$ are chosen the same as in Lemma \ref{choosek_single}.  Then the output $\yb_{\text{out}}$ of Algorithm \ref{algorithm:general} satisfies $\EE\|\dF(\yb_{\text{out}})\|_2^2 \leq \epsilon^2$
with less than
\begin{align}
    O\bigg(\log^3 \bigg(\frac{\cV}{\epsilon^2}\land n\bigg)\bigg[\frac{\cV}{\epsilon^2}\land n + \frac{L \Delta_F}{\epsilon^2} \bigg[\frac{\cV}{\epsilon^2}\land n\bigg]^{1/2}\bigg]\bigg)\label{maintheorem_100}
\end{align}
stochastic gradient computations, where $\Delta_F = F(\zb_0 ) - F^*$.
\end{theorem}
%\CC{since the base batch size is constant, we may replace H with B}
\begin{remark}
If we treat $\cV, L$ and $\Delta_F$ as constants, and assume $\epsilon \ll 1$, then \eqref{maintheorem_100} can be simplified to $\tO(\epsilon^{-3} \land n^{1/2}\epsilon^{-2})$. This gradient complexity is strictly better than $O(\epsilon^{-10/3} \land n^{2/3}\epsilon^{-2})$, which is achieved by SCSG \citep{lei2017non}. Specifically, when $n\lesssim 1/\epsilon^2$, our proposed $\algname$ is faster than SCSG by a factor of $n^{1/6}$; when $n\gtrsim 1/\epsilon^2$, $\algname$ is faster than SCSG by a factor of $\epsilon^{-1/3}$. Moreover, $\algname$ also outperforms Natasha 2 \citep{allen2017natasha} which attains $\tO(\epsilon^{-3.25})$ gradient complexity and needs the additional Hessian Lipschitz condition. %\CC{Natasha 2 needs Hessian Lipschitz}
\end{remark}

\subsection{Convergence of SNVRG-PL}
We now consider the case when $F$ is a $\tau$-gradient dominated function. In general, we are able to find an $\epsilon$-approximate global minimizer of $F$ instead of only an $\epsilon$-approximate stationary point. Algorithm \ref{algorithm:gradientdom} uses Algorithm \ref{algorithm:general} as a component.
\begin{theorem}\label{theorem_gradientdo}
Suppose that $F$ has averaged $L$-Lipschitz gradient and stochastic gradients with bounded variance $\sigma^2$, $F$ is a $\tau$-gradient dominated function. In Algorithm \ref{algorithm:gradientdom}, let the base batch size $\bbatch = n\land (4\con_1\tau \cV/\epsilon)$, the number of epochs for SNVRG $S = 1 \lor (2\con_1 \tau L/\bbatch^{1/2})$ and the number of epochs  $U =\log(2\Delta_F/\epsilon)$. The rest parameters $(K,M,\{B_l\},\{T_l\})$ are chosen as the same in Lemma \ref{choosek_single}. Then the output $\zb_{\text{out}}$ of Algorithm \ref{algorithm:gradientdom} satisfies $\EE\big[F(\zb_{\text{out}}) - F^*\big] \leq \epsilon$ within
\begin{align}\label{eq:grad_com_pl}
    O\bigg(\log^3\bigg(n\land \frac{\tau\cV}{\epsilon}\bigg)  \log \frac{\Delta_F}{\epsilon}\bigg[n\land \frac{\tau\cV}{\epsilon} + \tau L \bigg[n\land \frac{\tau\cV}{\epsilon}\bigg]^{1/2}\bigg]\bigg)
\end{align}
stochastic gradient computations, where $\Delta_F = F(\zb_0 ) - F^*$
\end{theorem}

\begin{remark}
If we treat $\sigma^2$, $L$ and $\Delta_F$ as constants, then the gradient complexity in \eqref{eq:grad_com_pl} turns into $\tO(n\land\tau\epsilon^{-1}+\tau(n\land\tau\epsilon^{-1})^{1/2})$. Compared with nonconvex SVRG \citep{reddi2016fast} which achieves $\tO(n+\tau n^{2/3})$ gradient complexity, our $\algname$-PL is strictly better than SVRG in terms of the first summand and is faster than SVRG at least by a factor of $n^{1/6}$ in terms of the second summand. Compared with a more general variant of SVRG, namely, the SCSG algorithm \citep{lei2017non}, which attains $\tO\big(n\wedge\tau\epsilon^{-1}+\tau (n\wedge \tau\epsilon^{-1})^{2/3}\big)$ gradient complexity, $\algname$-PL also outperforms it by a factor of $(n\land\tau\epsilon^{-1})^{1/6}$.

\end{remark}

%\CC{strong convexity}
If we further assume that $F$ is $\lambda$-strongly convex, then it is easy to verify that $F$ is also $1/(2\lambda)$-gradient dominated. As a direct consequence, we have the following corollary:
%\CC{there is no definition of strong convexity in section 3}
\begin{corollary}\label{coro:strong_convex}
Under the same conditions and parameter choices as Theorem \ref{theorem_gradientdo}. If we additionally assume that $F$ is $\lambda$-strongly convex, then Algorithm \ref{algorithm:gradientdom} will outputs an $\epsilon$-approximate global minimizer within 
\begin{align}
    \tO\bigg(n\land \frac{\lambda \cV}{\epsilon} + \kappa\cdot \bigg[n\land \frac{\lambda \cV}{\epsilon}\bigg]^{1/2}\bigg)\label{strongconvex}
\end{align}
stochastic gradient computations, where $\kappa = L/\lambda$ is the condition number of $F$.
\end{corollary}

\begin{remark}
Corollary \ref{coro:strong_convex} suggests that when we regard $\lambda$ and $\cV$ as constants and set $\epsilon \ll 1$, Algorithm \ref{algorithm:gradientdom} is able to find an $\epsilon$-approximate global minimizer within $\tO(n+ n^{1/2}\kappa)$ stochastic gradient computations, which matches SVRG-lep in Katyusha X \citep{allen2018katyusha}. Using catalyst techniques \citep{lin2015universal} or Katyusha momentum \citep{allen2017katyusha}, it can be further accelerated to $\tO(n+n^{3/4}\sqrt{\kappa})$, which matches the best-known convergence rate \citep{shalev2015sdca, allen2018katyusha}.
\end{remark}

\section{Experiments}\label{sec:experiment}

In this section, we compare our algorithm SNVRG with other baseline algorithms on training a convolutional neural network for image classification.  %\CC{add the experimental enviromental, pytorch, GPU, xxx}
%\subsection{Baseline Algorithms}
We compare the performance of the following algorithms: \emph{SGD};  SGD with momentum \citep{qian1999momentum} (denoted by \emph{SGD-momentum}); \emph{ADAM}\citep{kingma2014adam};  \emph{SCSG} \cite{lei2017non}. It is worth noting that \emph{SCSG} is a special case of SNVRG when the number of nested loops $K = 1$. Due to the memory cost, we did not compare \emph{GD} and \emph{SVRG} which need to calculate the full gradient. Although our theoretical analysis holds for general $K$ nested loops, it suffices to choose $K = 2$ in \emph{SNVRG} to illustrate the effectiveness of the nested structure for the simplification of implementation. In this case, we have $3$ reference points and gradients. 
%Although our theoretical analysis holds for general $K$ nested loops, here we choose $K = 2$ (thus we have $3$ reference points and gradients) for the simplification of implementation. 
%The details parameter tuning of these algorithms will be provided in each experiment. 
All experiments are conducted on Amazon AWS p2.xlarge servers which comes with Intel Xeon E5 CPU and NVIDIA Tesla K80 GPU (12G GPU RAM). All algorithm are implemented in Pytorch platform version $0.4.0$ within Python $3.6.4$.%\CC{where are parameter settings?}

\iffalse
\subsection{Nonconvex ERM on covtype dataset}
The first experiment is a binary logistic regression problem with a nonconvex regularizer. Given training data $\xb_i \in \RR^d$ and label $y_i \in \{0,1\}$, $1 \leq i \leq n$, the problem is formulated as follows
\begin{align}
    \min_{\sbb \in \RR^d}& \frac{1}{n}\sum_{i=1}^n\big[y_i\cdot  \log \phi(\sbb^{\top}\xb_i) + (1-y_i)\cdot \log [1- \phi(\sbb^{\top} \xb_i)]\big] + \lambda \sum_{i=1}^d \frac{(\alpha s_i)^2}{1+(\alpha s_i)^2}, 
\end{align}
where $\phi(x) = 1/(1+\exp(-x))$ is the sigmoid function. We fix $\lambda = 10^{-3}, \alpha = 1$ in this experiment. We choose \emph{covtype} dataset which comes from Libsvm \citep{chang2011libsvm}, where $n = 581,012$, $d = 54$. We use the loss function value gap and the gradient norm to compare different algorithms. For different algorithms, we tuned the their batch size using grid search to find the best batch size. After the tuning, for \emph{SGD} and \emph{SGD-momentum}, we use batch size equal to $1024$. For \emph{SVRG}, we use the batch sizes $(B,B_1)$ as $(581012,512)$. For \emph{SNVRG} we set $(B, B_1, B_2) = (581012, 512,16)$. The momemtum parameter for \emph{SGD-momentum} is set to $0.9$. The step size for all the compared algorithms are also carefully tuned by grid search. The experiment results were plotted in Figure \ref{fig:covtype_loss} and \ref{fig:covtype_gradnorm}. As we can see, our algorithm outperforms all the other baseline algorithms by a large margin. This is well-aligned with our theoretical analysis.
\fi

%\subsection{Convolutional neural network}

\noindent\textbf{Datasets} We use three image datasets: (1) The MNIST dataset \citep{scholkopf2002learning} consists of handwritten digits and has $50,000$ training examples and $10,000$ test examples. The digits have been size-normalized to fit the network, and each image is 28 pixels by 28 pixels. (2) CIFAR10 dataset \citep{krizhevsky2009learning} consists of images in 10 classes and has $50,000$ training examples and $10,000$ test examples. The digits have been size-normalized to fit the network, and each image is 32 pixels by 32 pixels. (3) SVHN dataset \cite{netzer2011reading} consists of images of digits and has $531,131$ training examples and $26,032$ test examples. The digits have been size-normalized to fit the network, and each image is 32 pixels by 32 pixels.

\noindent\textbf{CNN Architecture} We use the standard LeNet \citep{lecun1998gradient}, which has two convolutional layers with 6 and 16
filters of size 5 respectively, followed by three fully-connected layers with output size 120, 84 and 10. We apply max pooling after each convolutional layer.

\noindent\textbf{Implementation Details \& Parameter Tuning} 
% We compare \emph{SGD}, \emph{SGD-momentum}, \emph{ADAM}, \emph{SCSG} and \emph{SNVRG} in these three experiments. 
We did not use the random data augmentation which is set as default by Pytorch, because it will apply random transformation (e.g., clip and rotation) at the beginning of each epoch on the original image dataset, which will ruin the finite-sum structure of the loss function. We set our grid search rules for all three datasets as follows. For \emph{SGD}, we search the batch size from $\{256, 512, 1024, 2048\}$ and the initial step sizes from $\{1, 0.1, 0.01\}$. For \emph{SGD-momentum}, we set the momentum parameter as $0.9$. We search its batch size from $\{256, 512, 1024, 2048\}$ and the initial learning rate from $\{1, 0.1, 0.01\}$. For \emph{ADAM}, we search the batch size from $\{256, 512, 1024, 2048\}$ and the initial learning rate from $\{0.01, 0.001, 0.0001\}$. For \emph{SCSG} and \emph{SNVRG}, we choose loop parameters $\{T_l\}$ which satisfy $B_l\cdot \prod_{j=1}^l T_j = B $ automatically. In addition, for \emph{SCSG}, we set the batch sizes $(B, B_1) = (B, B/b)$, where $b$ is the batch size ratio parameter. We search $B$ from $\{256, 512, 1024, 2048\}$ and we search $b$ from $\{2,4,8\}$. We search its initial learning rate from $\{1, 0.1, 0.01\}$. For our proposed \emph{SNVRG}, we set the batch sizes $(B, B_1, B_2) = (B, B/b, B/b^2)$, where $b$ is the batch size ratio parameter. We search $B$ from $\{256, 512, 1024, 2048\}$ and $b$ from $\{2,4,8\}$. We search its initial learning rate from $\{1, 0.1, 0.01\}$. Following the convention of deep learning practice, we apply learning rate decay schedule to each algorithm with the learning rate decayed by $0.1$ every 20 epochs. We also conducted experiments based on plain implementation of different algorithms without learning rate decay, which is deferred to the appendix.

\iffalse
Due to the memory cost, we did not compare \emph{GD} and \emph{SVRG} because both of them need to calculate full gradient. Instead, we compare \emph{SGD}, \emph{SGD-momentum}, \emph{ADAM}, \emph{SCSG} and \emph{SNVRG} in these three experiments. For different algorithms, we tuned the their batch size using grid search to find the best batch size. After the tuning, for the MNIST dataset, we use \emph{SGD} with batch size $512$, \emph{SGD-momentum} with batch size $512$, \emph{SCSG} with batch size $(B, B_1)=(512,128)$, and \emph{SNVRG} with $(B, B_1, B_2) = (512, 128,8)$ in this experiment. For the Cifar10 dataset, we use \emph{SGD} with batch size $512$, \emph{SGD-momentum} with batch size $512$, \emph{SCSG} with batch size $(B, B_1)=(512,128)$, and \emph{SNVRG} with $(B, B_1, B_2) = (512, 128,4)$. For the SVHN dataset, we use \emph{SGD} with batch size $1024$, \emph{SGD-momentum} with batch size $1024$, \emph{SCSG} with batch size $(B, B_1)=(1024,256)$, and \emph{SNVRG} with $(B, B_1, B_2) = (1024, 256,8)$. The momentum parameter for \emph{SGD-momentum} is set to $0.9$ on all these 3 datasets. In addition, we choose the initial step sizes of \emph{SGD} and \emph{SGD-momentum} as $0.1$, and let them decay by $0.1$ every $20$ epochs following the convention of deep learning practice. For the step sizes of \emph{SCSG, SNVRG}, we choose them as constant and carefully tune them using grid search for the purpose of a fair comparison. 
\fi

\begin{figure}[t]
%\vskip -0.2in
	\begin{center}
		%\subfigure[training loss (\textit{covtype})]{\includegraphics[width=0.238\linewidth]{nips18/figures/covtype-epoch_loss.pdf}
		%\label{fig:covtype_loss}}
		\subfigure[training loss (\textit{MNIST})]{\includegraphics[width=0.32\linewidth]{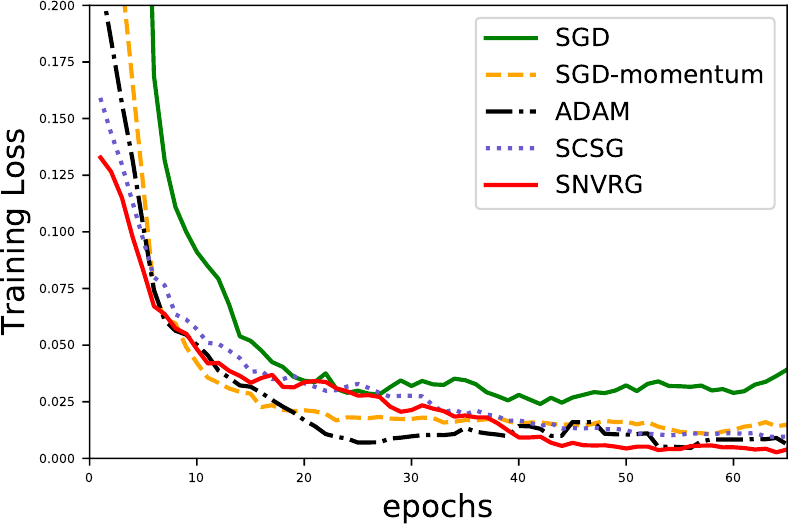}
		\label{fig:MNIST_loss}}		
		\subfigure[training loss (\textit{CIFAR10})]{\includegraphics[width=0.32\linewidth]{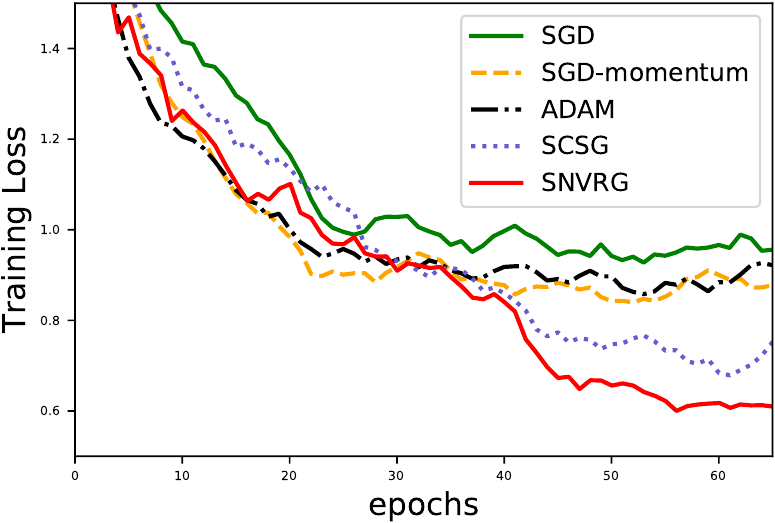}
		\label{fig:CIFAR10_loss}}
    	\subfigure[training loss (\textit{SVHN})]{\includegraphics[width=0.32\linewidth]{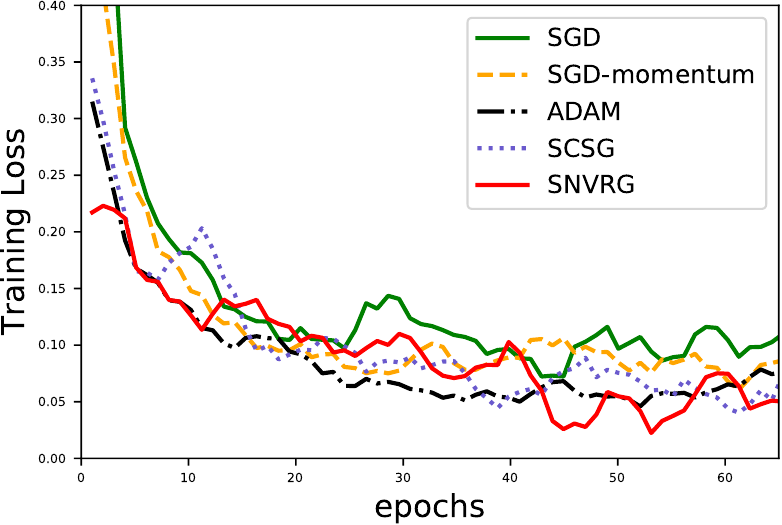}
    	\label{fig:svhn_loss}}		
		
		%\subfigure[grad. norm (\textit{covtype})]{\includegraphics[width=0.24\linewidth]{nips18/figures/covtype-epoch_gradnorm.pdf}\label{fig:covtype_gradnorm}}
		\subfigure[test error (\textit{MNIST})]{\includegraphics[width=0.32\linewidth]{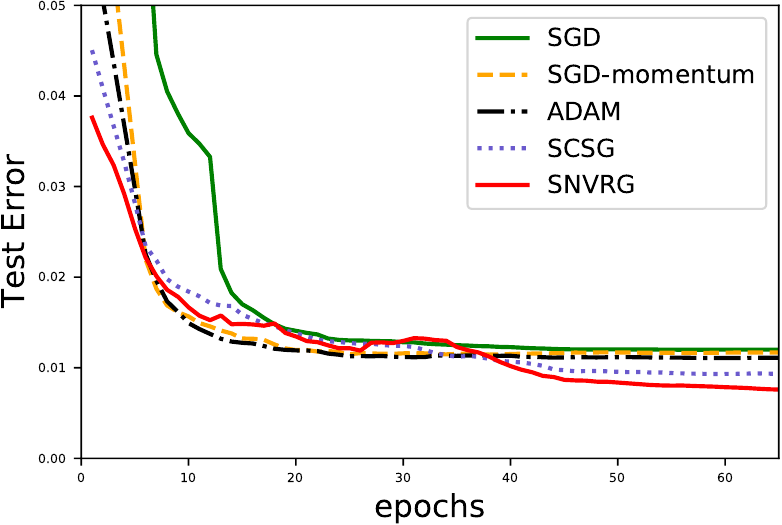}
		\label{fig:MNIST_validation}}		
		\subfigure[test error (\textit{CIFAR10})]{\includegraphics[width=0.32\linewidth]{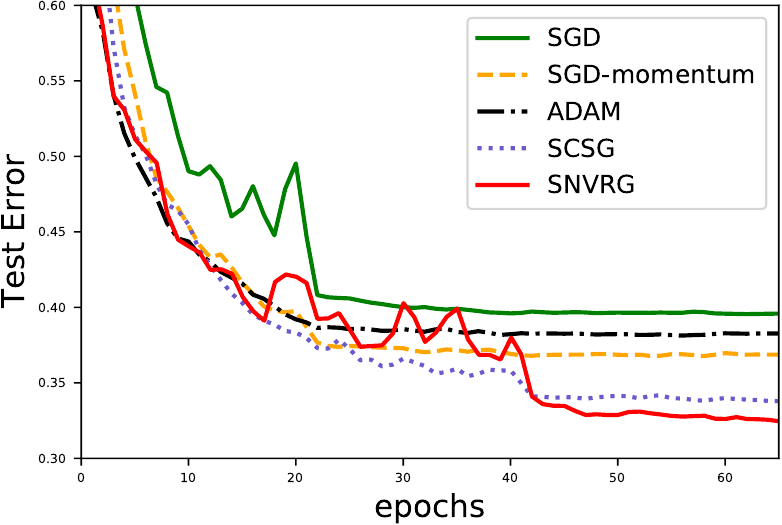}\label{fig:CIFAR10_validation}}
    	\subfigure[test error (\textit{SVHN})]{\includegraphics[width=0.32\linewidth]{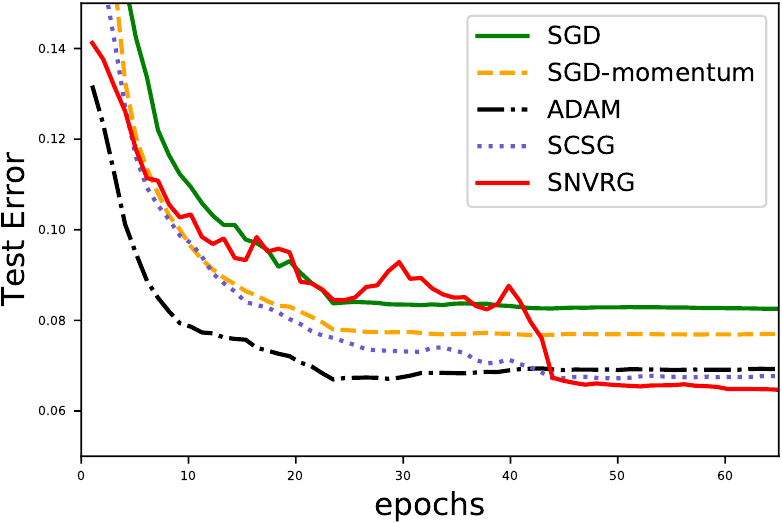}
    	\label{fig:svhn_validation}}
		
	\caption{Experiment results on different datasets with learning rate decay. 
	%(a) and (e) depict the function value gap and the gradient norm (grad. norm) v.s. data epochs for the binary logistic regression problem on \emph{covtype} dataset.
	(a) and (d) depict the training loss and test error (top-1 error) v.s. data epochs for training LeNet on MNIST dataset. (b) and (e) depict the training loss and test error v.s. data epochs for training LeNet on CIFAR10 dataset. (c) and (f) depict the training loss and test error v.s. data epochs for training LeNet on SVHN dataset. \label{fig:exp_result}}
	\end{center}
	%\vskip -0.1in
\end{figure}

\iffalse
We plotted the training loss and test error for each dataset in Figure \ref{fig:exp_result}. The results on MNIST are presented in Figures \ref{fig:MNIST_loss} and \ref{fig:MNIST_validation}; The results on CIFAR10 are in Figures \ref{fig:CIFAR10_loss} and \ref{fig:CIFAR10_validation}; The results on SVHN dataset are shown in Figures \ref{fig:svhn_loss} and \ref{fig:svhn_validation}. It can be seen that our algorithm \emph{SNVRG} outperforms all other baseline algorithms, which indicates that the use of nested reference points and gradients is able to accelerate the nonconvex finite-sum optimization. We can also observe that \emph{SGD-momentum} improves upon the \emph{SGD} algorithm during our experiment, but the improvement is not able to compete with the acceleration brought by variance reduction. We would like to emphasize that, while this experiment is on training convolutional neural networks, the major goal of this experiment is to illustrate the advantage of our algorithm and corroborate our theory, rather than claiming a state-of-the-art algorithm for training deep neural networks.
\fi

We plotted the training loss and test error for different algorithms on each dataset in Figure \ref{fig:exp_result}. The results on MNIST are presented in Figures \ref{fig:MNIST_loss} and \ref{fig:MNIST_validation}; the results on CIFAR10 are in Figures \ref{fig:CIFAR10_loss} and \ref{fig:CIFAR10_validation}; and the results on SVHN dataset are shown in Figures \ref{fig:svhn_loss} and \ref{fig:svhn_validation}. It can be seen that with learning rate decay schedule, our algorithm \emph{SNVRG} outperforms all baseline algorithms, which confirms that the use of nested reference points and gradients can accelerate the nonconvex finite-sum optimization. %We can observe that \emph{SGD-momentum} and $\emph{ADAM}$ also improves upon the \emph{SGD} algorithm during our experiment, but the improvement is not able to compete with the acceleration brought by variance reduction.

We would like to emphasize that, while this experiment is on training convolutional neural networks, the major goal of this experiment is to illustrate the advantage of our algorithm and corroborate our theory, rather than claiming a state-of-the-art algorithm for training deep neural networks.
%The second problem is a multi-classification problem which runs on a modified LeNet \citep{lecun1998gradient}, which has two convolutional layers with 10 and 20
%filters of size 5 respectively, followed by two fully-connected layers with output size 50 and 10. We apply max pooling after each convolutional layer. 

%\subsection{Convolutional Neural network on SVHN}
% We use 

%\subsection{Neural network on CIFAR10}
%The third problem is a multi-classification problem which runs on standard LeNet \citep{lecun1998gradient}, which has two convolutional layers with 6 and 16
%filters of size 5 respectively, followed by three fully-connected layers with output size 120, 84 and 10. We apply max pooling after each convolutional layer. 

\section{Conclusions and Future Work}\label{sec:conclusion}

In this paper, we proposed a stochastic nested variance reduced gradient method for finite-sum nonconvex optimization. It achieves substantially better gradient complexity than existing first-order algorithms. This partially resolves a long standing question that whether the dependence of gradient complexity on $n$ for nonconvex SVRG and SCSG can be further improved. There is still an open question: whether $\tilde O (n\land \epsilon^{-2}+\epsilon^{-3}\land n^{1/2}\epsilon^{-2})$ is the optimal gradient complexity for finite-sum and stochastic nonconvex optimization problem? 
%For finite sum convex optimization, the lower bound has been studied in a sequence of work \citep{agarwal2014lower, lan2017optimal, arjevani2016dimension, woodworth2016tight}. %\CC{also cite alekh aggarwal and guanghui lan} 
For finite-sum nonconvex optimization problem, the lower bound has been proved in \citet{fang2018spider}, which suggests that our algorithm is near optimal up to a logarithmic factor. However, for general stochastic problem, the lower bound is still unknown. We plan to derive such lower bound in our future work. On the other hand, our algorithm can also be extended to deal with nonconvex nonsmooth finite-sum optimization using proximal gradient \citep{reddi2016proximal}.

\section*{Acknowledgement}
We would like to thank the anonymous reviewers for their helpful comments. This research was sponsored in part by the National Science Foundation IIS-1652539 and BIGDATA IIS-1855099. We also thank AWS for providing cloud computing credits associated with the NSF BIGDATA award. The views and conclusions contained in this paper are those of the authors and should not be interpreted as representing any funding agencies.

\bibliographystyle{plainnat}
\bibliography{reference}

\begin{thebibliography}{48}
\providecommand{\natexlab}[1]{#1}
\providecommand{\url}[1]{\texttt{#1}}
\expandafter\ifx\csname urlstyle\endcsname\relax
  \providecommand{\doi}[1]{doi: #1}\else
  \providecommand{\doi}{doi: \begingroup \urlstyle{rm}\Url}\fi

\bibitem[Agarwal et~al.(2017)Agarwal, Allenzhu, Bullins, Hazan, and
  Ma]{Agarwal2017Finding}
Naman Agarwal, Zeyuan Allenzhu, Brian Bullins, Elad Hazan, and Tengyu Ma.
\newblock Finding approximate local minima for nonconvex optimization in linear
  time.
\newblock 2017.

\bibitem[Allen-Zhu(2017{\natexlab{a}})]{allen2017katyusha}
Zeyuan Allen-Zhu.
\newblock Katyusha: The first direct acceleration of stochastic gradient
  methods.
\newblock In \emph{Proceedings of the 49th Annual ACM SIGACT Symposium on
  Theory of Computing}, pages 1200--1205. ACM, 2017{\natexlab{a}}.

\bibitem[Allen-Zhu(2017{\natexlab{b}})]{allen2017natasha}
Zeyuan Allen-Zhu.
\newblock Natasha 2: Faster non-convex optimization than sgd.
\newblock \emph{arXiv preprint arXiv:1708.08694}, 2017{\natexlab{b}}.

\bibitem[Allen-Zhu(2018)]{allen2018katyusha}
Zeyuan Allen-Zhu.
\newblock {K}atyusha x: Simple momentum method for stochastic sum-of-nonconvex
  optimization.
\newblock In \emph{Proceedings of the 35th International Conference on Machine
  Learning}, volume~80, pages 179--185. PMLR, 10--15 Jul 2018.

\bibitem[Allen-Zhu and Hazan(2016)]{allen2016variance}
Zeyuan Allen-Zhu and Elad Hazan.
\newblock Variance reduction for faster non-convex optimization.
\newblock In \emph{International Conference on Machine Learning}, pages
  699--707, 2016.

\bibitem[Bietti and Mairal(2017)]{bietti2017stochastic}
Alberto Bietti and Julien Mairal.
\newblock Stochastic optimization with variance reduction for infinite datasets
  with finite sum structure.
\newblock In \emph{Advances in Neural Information Processing Systems}, pages
  1622--1632, 2017.

\bibitem[Carmon et~al.(2016)Carmon, Duchi, Hinder, and
  Sidford]{Carmon2016Accelerated}
Yair Carmon, John~C. Duchi, Oliver Hinder, and Aaron Sidford.
\newblock Accelerated methods for non-convex optimization.
\newblock 2016.

\bibitem[Carmon et~al.(2017{\natexlab{a}})Carmon, Duchi, Hinder, and
  Sidford]{carmon2017convex}
Yair Carmon, John~C Duchi, Oliver Hinder, and Aaron Sidford.
\newblock ``convex until proven guilty": Dimension-free acceleration of
  gradient descent on non-convex functions.
\newblock In \emph{International Conference on Machine Learning}, pages
  654--663, 2017{\natexlab{a}}.

\bibitem[Carmon et~al.(2017{\natexlab{b}})Carmon, Duchi, Hinder, and
  Sidford]{carmonlower}
Yair Carmon, John~C Duchi, Oliver Hinder, and Aaron Sidford.
\newblock Lower bounds for finding stationary points of non-convex, smooth
  high-dimensional functions.
\newblock 2017{\natexlab{b}}.

\bibitem[Defazio et~al.(2014{\natexlab{a}})Defazio, Bach, and
  Lacoste-Julien]{defazio2014saga}
Aaron Defazio, Francis Bach, and Simon Lacoste-Julien.
\newblock Saga: A fast incremental gradient method with support for
  non-strongly convex composite objectives.
\newblock In \emph{Advances in Neural Information Processing Systems}, pages
  1646--1654, 2014{\natexlab{a}}.

\bibitem[Defazio et~al.(2014{\natexlab{b}})Defazio, Domke,
  et~al.]{defazio2014finito}
Aaron Defazio, Justin Domke, et~al.
\newblock Finito: A faster, permutable incremental gradient method for big data
  problems.
\newblock In \emph{International Conference on Machine Learning}, pages
  1125--1133, 2014{\natexlab{b}}.

\bibitem[Fang et~al.(2018)Fang, Li, Lin, and Zhang]{fang2018spider}
Cong Fang, Chris~Junchi Li, Zhouchen Lin, and Tong Zhang.
\newblock Spider: Near-optimal non-convex optimization via stochastic
  path-integrated differential estimator.
\newblock In \emph{Advances in Neural Information Processing Systems}, pages
  686--696, 2018.

\bibitem[Garber and Hazan(2015)]{garber2015fast}
Dan Garber and Elad Hazan.
\newblock Fast and simple pca via convex optimization.
\newblock \emph{arXiv preprint arXiv:1509.05647}, 2015.

\bibitem[Ghadimi and Lan(2012)]{ghadimi2012optimal}
Saeed Ghadimi and Guanghui Lan.
\newblock Optimal stochastic approximation algorithms for strongly convex
  stochastic composite optimization i: A generic algorithmic framework.
\newblock \emph{SIAM Journal on Optimization}, 22\penalty0 (4):\penalty0
  1469--1492, 2012.

\bibitem[Ghadimi and Lan(2016)]{ghadimi2016accelerated}
Saeed Ghadimi and Guanghui Lan.
\newblock Accelerated gradient methods for nonconvex nonlinear and stochastic
  programming.
\newblock \emph{Mathematical Programming}, 156\penalty0 (1-2):\penalty0 59--99,
  2016.

\bibitem[Harikandeh et~al.(2015)Harikandeh, Ahmed, Virani, Schmidt,
  Kone{\v{c}}n{\`y}, and Sallinen]{harikandeh2015stopwasting}
Reza Harikandeh, Mohamed~Osama Ahmed, Alim Virani, Mark Schmidt, Jakub
  Kone{\v{c}}n{\`y}, and Scott Sallinen.
\newblock Stopwasting my gradients: Practical svrg.
\newblock In \emph{Advances in Neural Information Processing Systems}, pages
  2251--2259, 2015.

\bibitem[Hillar and Lim(2013)]{hillar2013most}
Christopher~J Hillar and Lek-Heng Lim.
\newblock Most tensor problems are np-hard.
\newblock \emph{Journal of the ACM (JACM)}, 60\penalty0 (6):\penalty0 45, 2013.

\bibitem[Hu et~al.(2009)Hu, Pan, and Kwok]{hu2009accelerated}
Chonghai Hu, Weike Pan, and James~T Kwok.
\newblock Accelerated gradient methods for stochastic optimization and online
  learning.
\newblock In \emph{Advances in Neural Information Processing Systems}, pages
  781--789, 2009.

\bibitem[Johnson and Zhang(2013)]{johnson2013accelerating}
Rie Johnson and Tong Zhang.
\newblock Accelerating stochastic gradient descent using predictive variance
  reduction.
\newblock In \emph{Advances in neural information processing systems}, pages
  315--323, 2013.

\bibitem[Karimi et~al.(2016)Karimi, Nutini, and Schmidt]{karimi2016linear}
Hamed Karimi, Julie Nutini, and Mark Schmidt.
\newblock Linear convergence of gradient and proximal-gradient methods under
  the polyak-{\l}ojasiewicz condition.
\newblock In \emph{Joint European Conference on Machine Learning and Knowledge
  Discovery in Databases}, pages 795--811. Springer, 2016.

\bibitem[Kingma and Ba(2014)]{kingma2014adam}
Diederik~P Kingma and Jimmy Ba.
\newblock Adam: A method for stochastic optimization.
\newblock \emph{arXiv preprint arXiv:1412.6980}, 2014.

\bibitem[Krizhevsky(2009)]{krizhevsky2009learning}
Alex Krizhevsky.
\newblock Learning multiple layers of features from tiny images.
\newblock 2009.

\bibitem[Lan(2012)]{lan2012optimal}
Guanghui Lan.
\newblock An optimal method for stochastic composite optimization.
\newblock \emph{Mathematical Programming}, 133\penalty0 (1):\penalty0 365--397,
  2012.

\bibitem[Lan and Zhou(2017)]{lan2017optimal}
Guanghui Lan and Yi~Zhou.
\newblock An optimal randomized incremental gradient method.
\newblock \emph{Mathematical programming}, pages 1--49, 2017.

\bibitem[LeCun et~al.(1998)LeCun, Bottou, Bengio, and
  Haffner]{lecun1998gradient}
Yann LeCun, L{\'e}on Bottou, Yoshua Bengio, and Patrick Haffner.
\newblock Gradient-based learning applied to document recognition.
\newblock \emph{Proceedings of the IEEE}, 86\penalty0 (11):\penalty0
  2278--2324, 1998.

\bibitem[Lei et~al.(2017)Lei, Ju, Chen, and Jordan]{lei2017non}
Lihua Lei, Cheng Ju, Jianbo Chen, and Michael~I Jordan.
\newblock Non-convex finite-sum optimization via scsg methods.
\newblock In \emph{Advances in Neural Information Processing Systems}, pages
  2345--2355, 2017.

\bibitem[Li and Lin(2015)]{li2015accelerated}
Huan Li and Zhouchen Lin.
\newblock Accelerated proximal gradient methods for nonconvex programming.
\newblock In \emph{Advances in neural information processing systems}, pages
  379--387, 2015.

\bibitem[Li et~al.(2017)Li, Zhou, Liang, and Varshney]{li2017convergence}
Qunwei Li, Yi~Zhou, Yingbin Liang, and Pramod~K Varshney.
\newblock Convergence analysis of proximal gradient with momentum for nonconvex
  optimization.
\newblock \emph{arXiv preprint arXiv:1705.04925}, 2017.

\bibitem[Lin et~al.(2015)Lin, Mairal, and Harchaoui]{lin2015universal}
Hongzhou Lin, Julien Mairal, and Zaid Harchaoui.
\newblock A universal catalyst for first-order optimization.
\newblock In \emph{Advances in Neural Information Processing Systems}, pages
  3384--3392, 2015.

\bibitem[Mairal(2015)]{mairal2015incremental}
Julien Mairal.
\newblock Incremental majorization-minimization optimization with application
  to large-scale machine learning.
\newblock \emph{SIAM Journal on Optimization}, 25\penalty0 (2):\penalty0
  829--855, 2015.

\bibitem[Nesterov(2005)]{nesterov2005smooth}
Yu~Nesterov.
\newblock Smooth minimization of non-smooth functions.
\newblock \emph{Mathematical programming}, 103\penalty0 (1):\penalty0 127--152,
  2005.

\bibitem[Nesterov(2014)]{Nesterov2014Introductory}
Yurii Nesterov.
\newblock \emph{Introductory Lectures on Convex Optimization}.
\newblock Kluwer Academic Publishers, 2014.

\bibitem[Netzer et~al.()Netzer, Wang, Coates, Bissacco, Wu, and
  Ng]{netzer2011reading}
Yuval Netzer, Tao Wang, Adam Coates, Alessandro Bissacco, Bo~Wu, and Andrew~Y
  Ng.
\newblock Reading digits in natural images with unsupervised feature learning.

\bibitem[Nguyen et~al.(2017{\natexlab{a}})Nguyen, Liu, Scheinberg, and
  Tak{\'a}{\v{c}}]{nguyen2017sarah}
Lam~M Nguyen, Jie Liu, Katya Scheinberg, and Martin Tak{\'a}{\v{c}}.
\newblock Sarah: A novel method for machine learning problems using stochastic
  recursive gradient.
\newblock In \emph{Proceedings of the 34th International Conference on Machine
  Learning-Volume 70}, pages 2613--2621. JMLR. org, 2017{\natexlab{a}}.

\bibitem[Nguyen et~al.(2017{\natexlab{b}})Nguyen, Liu, Scheinberg, and
  Tak{\'a}{\v{c}}]{nguyen2017stochastic}
Lam~M Nguyen, Jie Liu, Katya Scheinberg, and Martin Tak{\'a}{\v{c}}.
\newblock Stochastic recursive gradient algorithm for nonconvex optimization.
\newblock \emph{arXiv preprint arXiv:1705.07261}, 2017{\natexlab{b}}.

\bibitem[Paquette et~al.(2017)Paquette, Lin, Drusvyatskiy, Mairal, and
  Harchaoui]{paquette2017catalyst}
Courtney Paquette, Hongzhou Lin, Dmitriy Drusvyatskiy, Julien Mairal, and Zaid
  Harchaoui.
\newblock Catalyst acceleration for gradient-based non-convex optimization.
\newblock \emph{arXiv preprint arXiv:1703.10993}, 2017.

\bibitem[Polyak(1964)]{polyak1964some}
Boris~T Polyak.
\newblock Some methods of speeding up the convergence of iteration methods.
\newblock \emph{USSR Computational Mathematics and Mathematical Physics},
  4\penalty0 (5):\penalty0 1--17, 1964.

\bibitem[Polyak(1963)]{polyak1963gradient}
Boris~Teodorovich Polyak.
\newblock Gradient methods for minimizing functionals.
\newblock \emph{Zhurnal Vychislitel'noi Matematiki i Matematicheskoi Fiziki},
  3\penalty0 (4):\penalty0 643--653, 1963.

\bibitem[Qian(1999)]{qian1999momentum}
Ning Qian.
\newblock On the momentum term in gradient descent learning algorithms.
\newblock \emph{Neural networks}, 12\penalty0 (1):\penalty0 145--151, 1999.

\bibitem[Reddi et~al.(2016{\natexlab{a}})Reddi, Hefny, Sra, Poczos, and
  Smola]{Reddi2016Stochastic}
Sashank~J. Reddi, Ahmed Hefny, Suvrit Sra, Barnabas Poczos, and Alex Smola.
\newblock Stochastic variance reduction for nonconvex optimization.
\newblock pages 314--323, 2016{\natexlab{a}}.

\bibitem[Reddi et~al.(2016{\natexlab{b}})Reddi, Sra, P{\'o}czos, and
  Smola]{reddi2016fast}
Sashank~J Reddi, Suvrit Sra, Barnab{\'a}s P{\'o}czos, and Alex Smola.
\newblock Fast incremental method for smooth nonconvex optimization.
\newblock In \emph{Decision and Control (CDC), 2016 IEEE 55th Conference on},
  pages 1971--1977. IEEE, 2016{\natexlab{b}}.

\bibitem[Reddi et~al.(2016{\natexlab{c}})Reddi, Sra, Poczos, and
  Smola]{reddi2016proximal}
Sashank~J Reddi, Suvrit Sra, Barnabas Poczos, and Alexander~J Smola.
\newblock Proximal stochastic methods for nonsmooth nonconvex finite-sum
  optimization.
\newblock In \emph{Advances in Neural Information Processing Systems}, pages
  1145--1153, 2016{\natexlab{c}}.

\bibitem[Roux et~al.(2012)Roux, Schmidt, and Bach]{roux2012stochastic}
Nicolas~L Roux, Mark Schmidt, and Francis~R Bach.
\newblock A stochastic gradient method with an exponential convergence \_rate
  for finite training sets.
\newblock In \emph{Advances in Neural Information Processing Systems}, pages
  2663--2671, 2012.

\bibitem[Sch{\"o}lkopf and Smola(2002)]{scholkopf2002learning}
Bernhard Sch{\"o}lkopf and Alexander~J Smola.
\newblock \emph{Learning with kernels: support vector machines, regularization,
  optimization, and beyond}.
\newblock MIT press, 2002.

\bibitem[Shalev-Shwartz(2015)]{shalev2015sdca}
Shai Shalev-Shwartz.
\newblock Sdca without duality.
\newblock \emph{arXiv preprint arXiv:1502.06177}, 2015.

\bibitem[Shalev-Shwartz(2016)]{shalev2016sdca}
Shai Shalev-Shwartz.
\newblock Sdca without duality, regularization, and individual convexity.
\newblock In \emph{International Conference on Machine Learning}, pages
  747--754, 2016.

\bibitem[Shalev-Shwartz and Zhang(2013)]{shalev2013stochastic}
Shai Shalev-Shwartz and Tong Zhang.
\newblock Stochastic dual coordinate ascent methods for regularized loss
  minimization.
\newblock \emph{Journal of Machine Learning Research}, 14\penalty0
  (Feb):\penalty0 567--599, 2013.

\bibitem[Xiao and Zhang(2014)]{xiao2014proximal}
Lin Xiao and Tong Zhang.
\newblock A proximal stochastic gradient method with progressive variance
  reduction.
\newblock \emph{SIAM Journal on Optimization}, 24\penalty0 (4):\penalty0
  2057--2075, 2014.

\end{thebibliography}

\newpage
\appendix
\section{Proof of the Main Theoretical Results}
In this section, we provide the proofs of our main theories in Section \ref{sec:theory}.
\iffalse
\subsection{Proof of Lemma \ref{fixk_single}}
\begin{proof}[Proof of Lemma \ref{fixk_single}]
We can check that the choice of $M, \{T_i\}, \{B_i\}$ in Lemma \ref{fixk_single} satisfy the requirement of Lemma \ref{mainlemma}. Moreover, we have
\begin{align}
    T = \prod_{j=1}^K T_j = \baseb^{\frac{2^K-1}{2^{K+1}-1}}, \label{fixk_single_10}
\end{align}
Submit \eqref{fixk_single_10} into \eqref{maintheorem_1}, we have \eqref{fixk_single_1} hold.
Now we consider how many gradient we need. We have $S$ outer loop, in each loop, we need to compute $\gb_0$ once, which takes $\baseb$ gradient; for $\gb_i$, we need to compute $\gb_j$ for $\prod_{j=1}^i T_j$ times, and $2B_i$ gradient for each $\gb_i$. Then the total gradient complexity $\cT$ is
\begin{align}
    \cT &=  \baseb + 2\cdot\sum_{i=1}^K B_i \cdot \prod_{j=1}^i T_j.
\end{align}
From the choice of parameters, we can derive out the following equalities:
\begin{align}
    & \prod_{j=1}^i T_j = \baseb^{\frac{2^i-1}{2^{K+1}-1}},\\
    &B_i \cdot \prod_{j=1}^i T_j  = 6^{K-i+1}\baseb,\\
    &\sum_{i=1}^K B_i <2\cdot 6^K\cdot \baseb.
\end{align}
Then we have
\begin{align}
\notag\\
\cT& \leq \baseb+4\cdot 6^K\cdot \baseb\leq
5\cdot 6^K\cdot \baseb.\label{complex_1}
\end{align}
\end{proof}
\fi
%%%%%%%%%%%%%%%%%%%%%%%%%%%%%%%%%%%%%%%%%

\subsection{Proof of Lemma \ref{choosek_single}}
We first prove our key lemma on One-epoch-$\algname$. In order to prove Lemma \ref{choosek_single}, we need the following supporting lemma:
\begin{lemma}\label{mainlemma}
Let $T = \prod_{l=1}^K T_l$. If the step size and batch size parameters in Algorithm \ref{algorithm:1} satisfy $M\geq 6L$ and $B_l \geq 6^{K-l+1}(\prod_{s = l}^K T_s)^2$, then the output of Algorithm \ref{algorithm:1} satisfies
\begin{align}
    &\ES \|\dF(\xb_{\text{out}})\|_2^2\leq \con\bigg(\frac{M}{T}\cdot\EE\big[F(\xb_{0})- F(\xb_{T})\big]+ \frac{2\cV}{\baseb}\cdot \ind(\baseb < n) \bigg),\label{maintheorem_1}
\end{align}
where $\con = 100$ is a constant.
\end{lemma}
\begin{proof}[Proof of Lemma \ref{choosek_single}]
Note that $B=2^{2^K}$, we can easily check that the choice of $M, \{T_l\}, \{B_l\}$ in Lemma \ref{choosek_single} satisfies the assumption of Lemma \ref{mainlemma}. Moreover, we have
\begin{align}
    T = \prod_{l=1}^K T_l = \baseb^{1/2}. \label{choosek_single_10}
\end{align}
We now submit \eqref{choosek_single_10} into \eqref{maintheorem_1}, which immediately implies \eqref{choosek_single_1}.

Next we compute how many stochastic gradient computations we need in total after we run One-epoch-$\algname$ once. According to the update of reference gradients in Algorithm \ref{algorithm:1}, we only update $\gb_t^{(0)}$ once at the beginning of Algorithm \ref{algorithm:1} (Line~\ref{alg_line:update_g0}), which needs $\baseb$ stochastic gradient computations. For $\gb_t^{(l)}$, we only need to update it when $0 = (t \mod \prod_{j=l+1}^K T_j)$, and thus we need to sample $\gb_t^{(l)}$ for $T/\prod_{j=l+1}^K T_j = \prod_{j=1}^l T_j$ times. We need $2B_l$ stochastic gradient computations for each sampling procedure (Line~\ref{alg_line:update_grad} in Algorithm \ref{algorithm:1}). We use $\cT$ to represent the total number of stochastic gradient computations, then based on above arguments we have
\begin{align}\label{eq:cal_TotalT}
    \cT &=  \baseb + 2\sum_{l=1}^K B_l \cdot \prod_{j=1}^l T_j.
\end{align}
Now we calculate $\cT$ under the parameter choice of Lemma \ref{choosek_single}. Note that we can easily verify the following results:
\begin{align}\label{eq:cal_BT}
     \prod_{j=1}^l T_j &= 2^{2^{l-1}}=\baseb^{\frac{2^l}{2^{K+1}}},\
     B_1 \cdot \prod_{j=1}^1 T_j =  2\times 6^{K}\baseb,\ B_l \cdot \prod_{j=1}^l T_j  =6^{K-l+1}\baseb.
\end{align}
Submit \eqref{eq:cal_BT} into \eqref{eq:cal_TotalT} yields the following results:
\begin{align}
    \cT&= B+2\bigg(2\times 6^{K}\baseb+\sum_{l=2}^K 6^{K-l+1}\baseb\bigg)\notag\\
    &<B+6\times 6^K\baseb\notag \\
    & = B+6\times 6^{\log \log \baseb} \baseb\notag \\
    & <
    B+6\baseb\log^3 \baseb.
\end{align}
Therefore, the total gradient complexity $\cT$ is bounded as follows.
\begin{align}
\cT
& \leq \baseb+6 \baseb\log^3 \baseb \leq
7\baseb\log^3 \baseb.
\end{align}
\end{proof}
%%%%%%%%%%%%%%%%%%%%%%%%%%%%%%%%%%%%%%%%%%%%%%%%%%%%%%%%%%
\subsection{Proof of Theorem \ref{maintheorem}}
Now we prove our main theorem which spells out the gradient complexity of $\algname$.
\begin{proof}[Proof of Theorem \ref{maintheorem}]
By \eqref{choosek_single_1} we have
%\CC{in the following places, use $\ind$ rather than $I$ to denote indicator function}
\begin{align}
    &\ES \|\dF(\yb_s)\|_2^2\leq \con\bigg(\frac{L}{\bbatch^{1/2}}\cdot\EE\big[F(\zb_{s-1})- F(\zb_s)\big]+ \frac{\cV}{\bbatch}\cdot \ind(\bbatch < n) \bigg),\label{maintheorem_4}
\end{align}
where $\con = 600$. Taking summation for  \eqref{maintheorem_4} over $s$ from $1$ to $S$,  we have
\begin{align}
    \sum_{s=1}^S\ES \|\dF(\yb_s)\|_2^2&\leq \con\bigg(\frac{L}{\bbatch^{1/2}}\cdot\EE\big[F(\zb_{0})- F(\zb_S)\big]+ \frac{\cV}{\bbatch}\cdot \ind(\bbatch < n)\cdot S\bigg).\label{maintheorem_2}
\end{align}
Dividing both sides of \eqref{maintheorem_2} by $S$, we immediately obtain
\begin{align}
    \EE\|\dF(\yb_{\text{out}})\|_2^2 &\leq \con\bigg(\frac{L \EE\big[F(\zb_{0})- F^*\big]}{S\bbatch^{1/2}} + \frac{\cV}{\bbatch}\cdot \ind(\bbatch < n) \bigg)\label{maintheorem_2.33} \\
    &=\con\bigg(\frac{L\Delta_F}{S \bbatch^{1/2}} + \frac{\cV}{\bbatch}\cdot \ind(\bbatch < n) \bigg) ,\label{maintheorem_2.4}
\end{align}
where \eqref{maintheorem_2.33} holds because $F(\zb_S) \geq F^*$ and by the definition $\Delta_F=F(\zb_0)-F^*$. By the choice of parameters in Theorem \ref{maintheorem}, we have $\bbatch = n \land (2\con \cV/\epsilon^2), S = 1 \lor (2\con L \Delta_F/(\bbatch^{1/2}\epsilon^2))$, which implies
\begin{align}\label{maintheorem_2.5}
\begin{split}
    \ind(\bbatch < n)\cdot \cV/\bbatch &\leq \epsilon^2/(2\con), \quad\text{and}\quad
    L  \Delta_F/(S \bbatch^{1/2}) \leq \epsilon^2/(2\con).
\end{split}
\end{align}
Submitting \eqref{maintheorem_2.5} into \eqref{maintheorem_2.4}, we have $\EE\|\dF(\yb_{\text{out}})\|_2^2 \leq 2 \con \epsilon^2/ (2\con) = \epsilon^2$. By Lemma \ref{choosek_single_1}, we have that each One-epoch-SNVRG takes less than $7\bbatch \log^3 \bbatch$ stochastic gradient computations. Since we have total $S$ epochs, so the total gradient complexity of Algorithm \ref{algorithm:general} is less than
\begin{align}
    S\cdot 7 \bbatch\log^3 \bbatch  &\leq 7 \bbatch\log^3 \bbatch+ \frac{L \Delta_F}{\epsilon^2}\cdot 7 \bbatch^{1/2}\log^3 \bbatch \notag \\
    & = O\bigg(\log^3 \bigg(\frac{\cV}{\epsilon^2}\land n\bigg)\bigg[\frac{\cV}{\epsilon^2}\land n + \frac{L \Delta_F}{\epsilon^2} \bigg[\frac{\cV}{\epsilon^2}\land n\bigg]^{1/2}\bigg]\bigg),
\end{align}
which leads to the conclusion.
\end{proof}
\subsection{Proof of Theorem \ref{theorem_gradientdo}}
We then prove the main theorem on gradient complexity of $\algname$ under gradient dominance condition (Algorithm \ref{algorithm:gradientdom}).
\begin{proof}[Proof of Theorem \ref{theorem_gradientdo}]
Following the proof of Theorem \ref{maintheorem}, we obtain a similar inequality with \eqref{maintheorem_2.33}:
\begin{align}
     \EE\|\dF(\zb_{u+1})\|_2^2 \leq \con \bigg(\frac{L\EE [F(\zb_u) - F^*]}{S\bbatch^{1/2}} + \frac{\cV}{\bbatch}\cdot \ind(\bbatch < n) \bigg).\label{theorem_gradientdo_1}
\end{align}
Since $F$ is a $\tau$-gradient dominated function, we have $\EE\|\dF(\zb_{u+1})\|_2^2 \geq 1/\tau\cdot \EE[F(\zb_{u+1}) - F^*]$ by Definition \ref{def:gradientdo}. Plugging this inequality into \eqref{theorem_gradientdo_1} yields
\begin{align}
    \EE\big[F(\zb_{u+1}) - F^*\big] &\leq \frac{\con \tau L}{S \bbatch^{1/2}}\cdot\EE\big[F(\zb_{u}) - F^*\big]+\frac{\con\tau\cV}{\bbatch}\cdot \ind(\bbatch < n) \notag \\
    & \leq \frac{1}{2}
    \EE\big[F(\zb_{u}) - F^*\big]+ \frac{\epsilon}{4}\label{theorem_gradientdo_2},
\end{align}
where the second inequality holds due to the choice of parameters $\bbatch = n\land (4\con_1\tau \cV/\epsilon)$ and $S = 1 \lor (2\con_1 \tau L/\bbatch^{1/2})$ for Algorithm \ref{algorithm:gradientdom} in Theorem \ref{theorem_gradientdo}. By \eqref{theorem_gradientdo_2} we can derive
\begin{align}
    \EE\big[F(\zb_{u+1}) - F^*\big] -  \frac{\epsilon}{2} \leq \frac{1}{2}\bigg(\EE\big[F(\zb_{u}) - F^*\big] -  \frac{\epsilon}{2}\bigg),\notag
\end{align}
which immediately implies
\begin{align}
    \EE\big[F(\zb_{U}) - F^*\big] -  \frac{\epsilon}{2} \leq \frac{1}{2^U} \bigg(\Delta_F - \frac{\epsilon}{2}\bigg) \leq \frac{\Delta_F}{2^U}.\label{theorem_gradientdo_3}
\end{align}
Plugging the number of epochs $U =\log(2\Delta_F/\epsilon) $ into \eqref{theorem_gradientdo_3}, we obtain $\EE\big[F(\zb_{U}) - F^*\big] \leq \epsilon$. Note that each epoch of Algorithm \ref{algorithm:gradientdom} needs at most $S\cdot 7\bbatch \log^3 \bbatch $ stochastic gradient computations by Theorem \ref{maintheorem} and Algorithm \ref{algorithm:gradientdom} has $U$ epochs, which implies the total stochastic gradient complexity %is upper bounded by
\begin{align}
    U\cdot S\cdot 7 \bbatch \log^3 \bbatch
    & = O\bigg(\log^3\bigg(n\land \frac{\tau\cV}{\epsilon}\bigg)  \log \frac{\Delta_F}{\epsilon}\bigg[n\land \frac{\tau\cV}{\epsilon} + \tau L \bigg[n\land \frac{\tau\cV}{\epsilon}\bigg]^{1/2}\bigg]\bigg).
\end{align}
\end{proof}

\section{Proof of Key Lemma \ref{mainlemma}}\label{proof_of_mainlemma}
In this section, we focus on proving Lemma \ref{mainlemma} which plays a pivotal role in proving our main theorems. Let $M, \{T_i\}, \{B_i\}, \baseb$ be the parameters as defined in Algorithm \ref{algorithm:1}. We denote $T = \prod_{l=1}^K T_l$.
We define filtration $\cF_t = \sigma(\xb_0, \dots,\xb_t)$.
Let $\{\xb_t^{(l)}\},\{\gb_t^{(l)}\}$ be the reference points and reference gradients in Algorithm \ref{algorithm:1}. We define $\vb_t^{(l)}$ as
\begin{align}
    \vb_t^{(l)} := \sum_{j=0}^l \gb_t^{(j)},\quad\text{for }0 \leq l \leq K. \label{deftunv}
\end{align}
% \iffalse
% \begin{proposition}\label{gooddef}
% Given $0 \leq t< t' <T$, suppose $r$ is the max index where $\xb_{t}^{(r)} = \xb_{t'}^{(r)} $, then for any $0\leq l \leq r$, we have $\gb_{t}^{(l)} = \gb_{t'}^{(l)}$ and $\vb_{t}^{(l)} = \vb_{t'}^{(l)}$.
% \end{proposition}
% \fi
We first present the following definition and two technical lemmas for the purpose of our analysis.
\begin{definition}\label{defck}
We define constant series $\{c^{(s)}_{j}\}$ as the following. For each $s$, we define $c^{(s)}_{T_s}$ as
\begin{align}
    c^{(s)}_{T_s} &= \frac{M}{6^{K-s+1} \prod_{l=s}^K T_l}. \label{defck_1}
\end{align}
When $0 \leq j < T_s$, we define $c^{(s)}_{j}$ by induction:
\begin{align}
    c^{(s)}_{j} &= \bigg(1+\frac{1}{T_s}\bigg)c^{(s)}_{j+1}+\frac{3L^2}{M}\cdot\frac{\prod_{l=s+1}^K T_l}{B_s}.\label{defck_2}
\end{align}
\end{definition}

\begin{lemma}\label{muloop}
For any $p, s$, where $1\leq s \leq K$ and $0 \leq p \prod_{j=s}^K T_j<(p+1)\prod_{j=s}^K T_j \leq \prod_{j=1}^K T_j$, we define
\begin{align}
    &\start  = p\cdot \prod_{j=s}^K T_j,\ \fin = \start+\prod_{j=s}^K T_j, \notag
\end{align}
for simplification. Then we have the following results:
\begin{align}
    &\ES\bigg[ \sum_{j=\start}^{\fin-1}\frac{\|\dF(\xb_j)\|_2^2}{ 100M}+ F(\xb_{\fin})+c_{T_s}^{(s)}\cdot \|\xb_{\fin} - \xb_{\start}\|_2^2\big| \cF_{\start}\bigg] \notag \\ &\leq
    F(\xb_{\start})+ \frac{2}{M}\cdot \EE\big[\|\dF(\xb_{\start}) - \vb_{\start}\|_2^2 \big|\cF_{\start}\big]\cdot \prod_{j=s}^K T_j. \notag
\end{align}
\end{lemma}

\begin{lemma}[\citet{lei2017non}]\label{concen}
Let $\ab_i$ be vectors satisfying $\sum_{i=1}^N \ab_i = 0$. Let $\cJ$ be a uniform random subset of $\{1,\dots,N\}$ with size $m$, then
\begin{align}
    \EE\bigg\|\frac{1}{m}\sum_{j \in \cJ}\ab_j\bigg\|_2^2 \leq \frac{\ind(|\cJ|<N)}{mN}\sum_{j=1}^N\|\ab_j\|_2^2.\notag
\end{align}
\end{lemma}

\begin{proof}[Proof of Lemma \ref{mainlemma}]
We have
\begin{align}
    \sum_{j=0}^{T-1}\frac{\ES \|\dF(\xb_j)\|_2^2}{100M}+ \EE\big[F(\xb_{T})\big]
    & \leq
    \sum_{j=0}^{T-1}\frac{\ES \|\dF(\xb_j)\|_2^2}{100M}+ \EE\big[F(\xb_{T})+c_{T_1}^{(1)}\cdot \|\xb_{T} - \xb_0\|_2^2\big] \notag\\
    & \leq
    \EE\big[F(\xb_0)\big]+ \frac{2}{M}\cdot \EE\|\dF(\xb_0) - \gb_0\|_2^2 \cdot T,\label{mainlemma_1}
\end{align}
where the second inequality comes from Lemma \ref{muloop} with we take $s = 1, p = 0$. Moreover we have
\begin{align}
    \EE\|\dF(\xb_0) - \gb_0\|_2^2  &= \EE\bigg\| \frac{1}{\baseb}\sum_{i\in I}\big[\df_i(\xb_0) - \dF(\xb_0)\big]\bigg\|_2^2\notag \\
    &\leq  \ind(\baseb<n)\cdot\frac{1}{\baseb}\cdot \frac{1}{n}\sum_{i=1}^n \big\|\df_i(\xb_0) - \dF(\xb_0)\big\|_2^2\label{mainlemma_20} \\
     & \leq \ind(\baseb<n)\cdot\frac{\cV}{\baseb},\label{mainlemma_2}
\end{align}
where \eqref{mainlemma_20} holds because of Lemma \ref{concen}. Plug \eqref{mainlemma_2} into \eqref{mainlemma_1} and note that we have $M=6 L$, and then we obtain
\begin{align}
    &\sum_{j=0}^{T-1}\ES \|\dF(\xb_j)\|_2^2\leq \con \bigg(M\EE\big[F(\xb_0)-F(\xb_{T})\big]+ \frac{2T\cV}{\baseb}\cdot \ind(\baseb < n) \bigg),\label{mainlemma_3}
\end{align}
where $\con = 100$. Divide both sides of \eqref{mainlemma_3} by $T$, then Lemma \ref{mainlemma} holds trivially.
\end{proof}

\section{Proof of Technical Lemmas}\label{app:technical lemmas}
In this section, we provide the proofs of technical lemmas used in Appendix \ref{proof_of_mainlemma}.
\subsection{Proof of Lemma \ref{muloop}}
Let $M, \{T_l\}, \{B_l\}, \baseb$ be the parameters defined in Algorithm \ref{algorithm:1} and $\{\xb_t^{(l)}\},\{\gb_t^{(l)}\}$ be the reference points and reference gradients defined in Algorithm \ref{algorithm:1}. Let $\vb_t^{(l)}, \cF_t$ be the variables and filtration defined in Appendix \ref{proof_of_mainlemma} and let $c_j^{(s)}$ be the constant series defined in Definition \ref{defck}.

In order to prove Lemma \ref{muloop}, we will need the following supporting propositions and lemmas. We first state the proposition about the relationship among $\xb_{t}^{(s)}, \gb_{t}^{(s)}$ and $\vb_{t}^{(s)}$:
\begin{proposition}\label{gooddef}
Let $\vb_t^{(l)}$ be defined as in \eqref{deftunv}. Let $p,s$ satisfy $0 \leq p\cdot \prod_{j=s+1}^K T_j < (p+1)\cdot \prod_{j=s+1}^K T_j < T$. For any $t, t'$ satisfying  $p\cdot \prod_{j=s+1}^K T_j \leq t <t'< (p+1)\cdot \prod_{j=s+1}^K T_j$, it holds that
\begin{align}
    &\xb_{t}^{(s)} = \xb_{t'}^{(s)} = \xb_{p \prod_{j=s+1}^K T_j}, \label{gooddef_1}\\
    &\gb_{t}^{(s')} = \gb_{t'}^{(s')},&\text{for any }s'\text{ that satisfies } 0 \leq s' \leq s,\label{gooddef_2}\\
    &\vb_{t}^{(s)} =\vb_{t'}^{(s)} =  \vb_{p \prod_{j=s+1}^K T_j}.\label{gooddef_3}
\end{align}
\end{proposition}

The following lemma spells out the relationship between $c_j^{(s-1)}$ and $c^{(s)}_{T_s}$. In a word, $c_j^{(s-1)}$ is about $1+T_{s-1}$ times less than $c^{(s)}_{T_s}$:
\begin{lemma}\label{propck}
If $B_s \geq 6^{K-s+1}(\prod_{l=s}^K T_l)^2, T_l\geq 1$ and $M \geq 6 L$, then it holds that
\begin{align}
    &c_j^{(s-1)}\cdot (1+T_{s-1}) < c^{(s)}_{T_s}, \qquad\text{for }2 \leq s \leq K, 0 \leq j \leq T_{s-1}, \label{propck_1}
\end{align}
and
\begin{align}
    c_{j}^{(K)}\cdot(1+T_K)<M, \qquad\text{for }0 \leq j \leq T_{K}. \label{propck_2}
\end{align}
% \begin{align}
%     \begin{cases}
%     c_j^{(s-1)}\cdot (1+T_{s-1}) < c^{(s)}_{T_s}, &\text{for }2 \leq s \leq K, 0 \leq j \leq T_{s-1}, \label{propck_1}\\
%     c_{j}^{(K)}\cdot(1+T_K)<M, &\text{for }0 \leq j \leq T_{K}. \label{propck_2}
%     \end{cases}
% \end{align}
\end{lemma}

Next lemma is a special case of Lemma \ref{muloop} with $s = K$:
\begin{lemma}\label{oneloop}
Suppose $p$ satisfies $0 \leq p T_K < (p+1) T_K \leq \prod_{i=1}^K T_i$. If $M>L$, then we have
\begin{align}
    &\ES \Bigg[F\big(\xb_{(p+1)\cdot T_K}\big)+c_{T_K}^{(K)}\cdot \big\|\xb_{(p+1)\cdot T_K} - \xb_{p\cdot T_K}\big\|_2^2+\sum_{j=0}^{T_{K}-1} \frac{\big\|\dF(\xb_{p\cdot T_K+j})\big\|_2^2}{100M}\bigg| \cF_{p\cdot T_K}\Bigg]\notag\\
    &\leq F(\xb_{p\cdot T_K})+\frac{2}{M}\cdot\EE\big[ \big\|\dF(\xb_{p\cdot T_K}) - \vb_{p\cdot T_K}\big\|_2^2\big|\cF_{p\cdot T_K}\big]\cdot T_K.
\end{align}
\end{lemma}

The following lemma provides an upper bound of $\EE \big[\big\|\dF(\xb_t^{(l)}) - \vb_t^{(l)}\big\|_2^2 \big]$, which plays an important role in our proof of Lemma \ref{muloop}.
\begin{lemma}\label{inductgra}
Let $t^l$ be as defined in \eqref{referencepoints}, then we have $\xb_t^{(l)} = \xb_{t^l}$, and
\begin{align}
    \EE \big[\big\|\dF(\xb_t^{(l)}) - \vb_t^{(l)}\big\|_2^2 \big| \cF_{t^l}\big]\leq \frac{L^2}{B_l}\big\|\xb_t^{(l)} - \xb_t^{(l-1)}\big\|_2^2 + \big\|\dF(\xb_t^{(l-1)}) - \vb_t^{(l-1)}\big\|_2^2.
\end{align}
\end{lemma}

\begin{proof}[Proof of Lemma \ref{muloop}]
We use mathematical induction to prove that Lemma \ref{muloop} holds for any $1 \leq s \leq K$. When $s = K$, the statement holds because of Lemma \ref{oneloop}. Suppose that for $s +1$, Lemma \ref{muloop} holds for any $p'$ which satisfies $0 \leq p' \prod_{j=s+1}^K T_j<(p'+1)\prod_{j=s+1}^K T_j \leq \prod_{j=1}^K T_j$. We need to prove Lemma \ref{muloop} still holds for $s$ and $p$, where $p$ satisfies $0 \leq p \prod_{j=s}^K T_j<(p+1)\prod_{j=s}^K T_j \leq \prod_{j=1}^K T_j$. We first define $m = \prod_{j=s+1}^K T_j$ for simplification, then we choose $p' = p T_s+u$, and we set indices $\start_u$ and $\fin_u$ as
\begin{align}
    &\start_u =  p' \prod_{j=s+1}^K T_j,\qquad \fin_u = \start_u+\prod_{j=s+1}^K T_j. \notag
\end{align}
It can be easily verified that the following relationship also holds:
\begin{align}\label{eq:start_fin_indice}
   \start_u = \start+u m, \qquad\fin_u = \start+(u+1) m .
\end{align}
Based on \eqref{eq:start_fin_indice}, we have
\begin{align}
&\EE\bigg[ \sum_{j=\start_u}^{\fin_u-1}\frac{\|\dF(\xb_j)\|_2^2}{100M}+ F(\xb_{\start+(u+1) m})+c_{T_{s+1}}^{(s+1)}\cdot \|\xb_{\start+(u+1) m} - \xb_{\start+u m}\|_2^2\big| \cF_{\start_u}\bigg] \notag\\
&=\EE\bigg[\sum_{j=\start_u}^{\fin_u-1}\frac{ \|\dF(\xb_j)\|_2^2}{100M}+ F(\xb_{\fin_u})+c_{T_{s+1}}^{(s+1)}\cdot \|\xb_{\fin_u} - \xb_{\start_u}\|_2^2\big| \cF_{\start_u}\bigg] \notag\\
& \leq F(\xb_{\start_u})+ \frac{2}{M}\cdot \EE\big[\big\|\dF(\xb_{\start_u}) - \vb_{\start_u}\big\|_2^2 \big|\cF_{\start_u}\big]\cdot \prod_{j=s+1}^K T_{j},\label{muloop_1}
\end{align}
where the last inequality holds because of the induction hypothesis that Lemma \ref{muloop} holds for $s+1$ and $p'$.
Note that we have $\xb_{\start_u} = \xb_{\start+u\cdot m} =\xb_{\start_u}^{(s)}$ from Proposition \ref{gooddef}, which implies
\begin{align}
    \EE\big[\big\|\dF(\xb_{\start_u}) - \vb_{\start_u}\big\|_2^2 \big|\cF_{\start_u}\big]
    & = \EE\big[\big\|\dF(\xb_{\start_u}^{(s)}) - \vb_{\start_u}^{(s)}\big\|_2^2 \big|\cF_{\start_u}\big] \notag  \\
    & \leq
    \frac{L^2}{B_s}\big\|\xb_{\start_u}^{(s)} -\xb_{\start_u}^{(s-1)} \big\|_2^2 + \big\|\dF(\xb_{\start_u}^{(s-1)}) - \vb_{\start_u}^{(s-1)}\big\|_2^2\label{muloop_1.33} \\
    & =
    \frac{L^2}{B_s}\|\xb_{\start+u\cdot m} - \xb_{\start}\|_2^2 + \big\|\dF(\xb_{\start}) - \vb_{\start}\big\|_2^2,\label{muloop_1.35}
\end{align}
where \eqref{muloop_1.33} holds because of Lemma \ref{inductgra} and \eqref{muloop_1.35} holds due to Proposition \ref{gooddef}. Plugging \eqref{muloop_1.35} into \eqref{muloop_1} and taking expectation $\EE[\cdot|\cF_{\start}]$ for \eqref{muloop_1}, we have
\begin{align}
     &\ES\bigg[\sum_{j=\start_u}^{\fin_u-1}\frac{ \|\dF(\xb_j)\|_2^2}{100M}+ F(\xb_{\start+(u+1) m})+c_{T_{s+1}}^{(s+1)} \|\xb_{\start+(u+1) m} - \xb_{\start+u m}\|_2^2\big|\cF_{\start}\bigg] \notag\\
    & \leq
    \EE\bigg[F(\xb_{\start+u m})+   \frac{2L^2}{MB_s}\|\xb_{\start+u m} - \xb_{\start}\|_2^2 \cdot \prod_{j=s+1}^K T_{j} +\frac{2}{M} \big\|\dF(\xb_{\start}) - \vb_{\start}\big\|_2^2 \cdot \prod_{j=s+1}^K T_{j} \bigg|\cF_{\start}\bigg].\label{muloop_3}
\end{align}
Next we bound $\|\xb_{\start+(u+1)\cdot m} - \xb_{\start}\|_2^2$ as the following:
\begin{align}
    &\|\xb_{\start+(u+1)\cdot m} - \xb_{\start}\|_2^2\notag\\
    & = \|\xb_{\start+u\cdot m} - \xb_{\start}\|_2^2 + \|\xb_{\start+(u+1)\cdot m} - \xb_{\start+u\cdot m}\|_2^2 \notag \\
    &\quad\quad + 2\la \xb_{\start+(u+1)\cdot m} - \xb_{\start+u\cdot m}, \xb_{\start+u\cdot m} - \xb_{\start} \ra \notag\\
    & \leq
    \|\xb_{\start+u\cdot m} - \xb_{\start}\|_2^2 + \|\xb_{\start+(u+1)\cdot m} - \xb_{\start+u\cdot m}\|_2^2 \notag \\
    &\quad\quad + \frac{1}{T_s}\cdot\|\xb_{\start+u\cdot m} - \xb_{\start}\|_2^2 + T_s\cdot\|\xb_{\start+(u+1)\cdot m} - \xb_{\start+u\cdot m}\|_2^2  \label{muloop_4}\\
    & =
    \bigg(1+\frac{1}{T_s}\bigg)\cdot\|\xb_{\start+u\cdot m} - \xb_{\start}\|_2^2 + (1+T_s)\cdot\|\xb_{\start+(u+1)\cdot m} - \xb_{\start+u\cdot m}\|_2^2 ,\label{muloop_5}
\end{align}
where \eqref{muloop_4} holds because of Young's inequality. Taking expectation $\EE[\cdot|\cF_{\start}]$ over \eqref{muloop_5} and multiplying $c_{u+1}^{(s)}$ on both sides, we obtain
\begin{align}
    c_{u+1}^{(s)}\EE\big[\|\xb_{\start+(u+1)\cdot m} - \xb_{\start}\|_2^2\big|\cF_{\start}\big]
    & \leq c_{u+1}^{(s)} \bigg(1+\frac{1}{T_s}\bigg)\EE\big[\|\xb_{\start+u\cdot m} - \xb_{\start}\|_2^2 \big|\cF_{\start}\big]\notag \\
    &\quad\quad + c_{u+1}^{(s)} (1+T_s)\EE\big[\|\xb_{\start+(u+1) m} - \xb_{\start+u m}\|_2^2\big|\cF_{\start}\big].\label{muloop_5.5}
\end{align}
Adding up inequalities\eqref{muloop_5.5} and \eqref{muloop_3} together yields
\begin{align}
    &\ES\bigg[\sum_{j=\start_u}^{\fin_u-1} \frac{\|\dF(\xb_j)\|_2^2}{100M}+ F(\xb_{\start+(u+1) m})+c_{u+1}^{(s)}\|\xb_{\start+(u+1) m} - \xb_{\start}\|_2^2\notag \\
    &\quad\quad +c_{T_{s+1}}^{(s+1)} \|\xb_{\start+(u+1) m} - \xb_{\start+u m}\|_2^2\big|\cF_{\start}\bigg] \notag\\
    & \leq
    \EE\bigg[F(\xb_{\start+u m})   + \|\xb_{\start+u m} - \xb_{\start}\|_2^2\bigg[c_{u+1}^{(s)}\bigg(1+\frac{1}{T_s}\bigg)+\frac{3L^2}{B_s M}
     \prod_{j=s+1}^K T_{j} \bigg]\bigg|\cF_{\start} \bigg]\notag\\
    &\quad\quad+
    \frac{2}{M}\EE\big[ \big\|\dF(\xb_{\start}) - \vb_{\start}\big\|_2^2\big|\cF_{\start}\big]  \prod_{j=s+1}^K T_{j} \notag\\
    &\qquad+c_{u+1}^{(s)}(1+T_s) \EE\big[\|\xb_{\start+(u+1) m} - \xb_{\start+u m}\|_2^2\big|\cF_{\start}\big]\notag\\
    & <
    \EE\big[F(\xb_{\start+u m})   + c_u^{(s)}\|\xb_{\start+u m} - \xb_{\start}\|_2^2\big|\cF_{\start} \big] +
    \frac{2}{M}  \EE\big[\big\|\dF(\xb_{\start}) - \vb_{\start}\big\|_2^2\big|\cF_{\start}\big]  \prod_{j=s+1}^K T_{j}\notag \\
    &\quad\quad+c_{T_{s+1}}^{(s+1)} \EE\big[\|\xb_{\start+(u+1) m} - \xb_{\start+u m}\|_2^2\big|\cF_{\start}\big],\label{muloop_6}
\end{align}
where the last inequality holds due to the fact that $c_u^{(s)} = c_{u+1}^{(s)}(1+1/T_s)+3L^2/(B_s M)\cdot \prod_{j=s+1}^K T_{j}$ by Definition \ref{defck} and $c_{u+1}^{(s)}\cdot(1+T_s)<c_{T_{s+1}}^{(s+1)}$ by Lemma \ref{propck}. Cancelling out the term $c_{T_{s+1}}^{(s+1)}\cdot \EE\big[\|\xb_{\start+(u+1)\cdot m} - \xb_{\start+u\cdot m}\|_2^2\big|\cF_{\start}\big]$ from both sides of \eqref{muloop_6}, we get
\begin{align}
    &\sum_{j=\start_u}^{\fin_u-1}\ES\bigg[\frac{ \|\dF(\xb_j)\|_2^2} {100M}\bigg|\cF_{\start}\bigg]+ \EE\big[F(\xb_{\start+(u+1)\cdot m})+c_{u+1}^{(s)}\cdot\|\xb_{\start+(u+1)\cdot m} - \xb_{\start}\|_2^2\big|\cF_{\start}\big] \notag\\
    & \leq
    \EE\big[F(\xb_{\start+u m})   + c_u^{(s)}\|\xb_{\start+u m} - \xb_{\start}\|_2^2 \big|\cF_{\start}\big] \notag\\
    &\qquad+\frac{2}{M} \EE\big[ \big\|\dF(\xb_{\start}) - \vb_{\start}\big\|_2^2 \big|\cF_{\start}\big] \prod_{j=s+1}^K T_{j}.\notag%\label{muloop_7}
\end{align}
We now telescope the above inequality for $u=0 $ to $T_s-1 $, then we have
\begin{align*}
    &  \ES \bigg[\sum_{u=0}^{T_s-1}\sum_{j=\start_u}^{\fin_u-1}\frac{\|\dF(\xb_j)\|_2^2}{100M}+ F(\xb_{\fin})+c_{T_s}^{(s)}\cdot \|\xb_{\fin} - \xb_{\start}\|_2^2\big|\cF_{\start}\bigg] \notag\\
    & \leq
    F(\xb_{\start})+ \frac{2T_s }{M}\cdot \EE\big[\big\|\dF(\xb_{\start}) - \vb_{\start}\big\|_2^2 \big|\cF_{\start}\big]\cdot \prod_{j=s+1}^K T_j .
\end{align*}
Since $\start_u = \fin_{u-1}, \start_0 = \start$, and $\fin_{T_s-1} = \fin$, we have
\begin{align}
    &\ES \bigg[\sum_{j=\start}^{\fin-1}\frac{\|\dF(\xb_j)\|_2^2}{100M}+ F(\xb_{\fin})+c_{T_s}^{(s)}\cdot \|\xb_{\fin} - \xb_{\start}\|_2^2\big|\cF_{\start}\bigg] \notag\\
   % & = \sum_{u=0}^{T_s-1}\sum_{j=\start_u}^{\fin_u-1}\ES \big[\|\dF(\xb_j)\|_2^2/(100M)+ F(\xb_{\fin})+c_{T_s}^{(s)}\cdot \|\xb_{\fin} - \xb_{\start}\|_2^2\big|\cF_{\start}\big] \notag\\
    & \leq
    F(\xb_{\start})+ \frac{2}{M}\cdot \EE\big[\big\|\dF(\xb_{\start}) - \vb_{\start}\big\|_2^2 \big|\cF_{\start}\big]\cdot \prod_{j=s}^K T_j.\label{muloop_8}
\end{align}
Therefore, we have proved that Lemma \ref{muloop} still holds for $s$ and $p$. Then by mathematical induction, we have for all $1\leq s \leq K$ and $p$ which satisfy $0 \leq p\cdot \prod_{j=s}^K T_j<(p+1)\cdot\prod_{j=s}^K T_j \leq \prod_{j=1}^K T_j$, Lemma \ref{muloop} holds.
\end{proof}

%very single loop

\subsection{Proof of Lemma \ref{concen}}
The following proof is adapted from that of Lemma A.1 in \citet{lei2017non}. We provide the proof here for the self-containedness of our paper.
\begin{proof}[Proof of Lemma \ref{concen}]
We only consider the case when $m<N$. Let $W_j = \ind(j \in \cJ)$, then we have
\begin{align}
    \EE W_j^2 = \EE W_j = \frac{m}{N}, \EE W_jW_{j'} = \frac{m(m-1)}{N(N-1)}.
\end{align}
Thus we can rewrite the sample mean as
\begin{align}
    \frac{1}{m}\sum_{j \in \cJ}\ab_j = \frac{1}{m}\sum_{i=1}^N W_i\ab_i,
\end{align}
which immediately implies
\begin{align}
    \EE\bigg\|\frac{1}{m}\sum_{j \in \cJ}\ab_j\bigg\|^2 &= \frac{1}{m^2}\bigg(\sum_{j=1}^N \EE W_j^2\|\ab_j\|_2^2+\sum_{j \neq j'}\EE W_jW_{j'}\la\ab_j, \ab_{j'}\ra\bigg)\notag \\
    & = \frac{1}{m^2}\bigg(\frac{m}{N}\sum_{j=1}^N\|\ab_j\|_2^2+\frac{m(m-1)}{N(N-1)}\sum_{j \neq j'}\la\ab_j, \ab_{j'}\ra \notag \\
    &=
    \frac{1}{m^2}\Bigg(\bigg(\frac{m}{N} - \frac{m(m-1)}{N(N-1)}\bigg)\sum_{j=1}^N\|\ab_j\|_2^2+\frac{m(m-1)}{N(N-1)}\bigg\|\sum_{j=1}^N\ab_j\bigg\|_2^2\Bigg)\notag \\
    & =
    \frac{1}{m^2}\bigg(\frac{m}{N} - \frac{m(m-1)}{N(N-1)}\bigg)\sum_{j=1}^N\|\ab_j\|_2^2\notag \\
    & \leq
    \frac{1}{m}\cdot \frac{1}{N}\sum_{j=1}^N\|\ab_j\|_2^2.\notag
\end{align}
\end{proof}

\section{Proofs of the Auxiliary Lemmas}
In this section, we present the additional proofs of supporting lemmas used in Appendix \ref{app:technical lemmas}. Let $M, \{T_l\}, \{B_l\}$ and $ \baseb$ be the parameters defined in Algorithm \ref{algorithm:1}. Let $\{\xb_t^{(l)}\},\{\gb_t^{(l)}\}$ be the reference points and reference gradients used in Algorithm \ref{algorithm:1}. Finally, $\vb_t^{(l)}, \cF_t$ are the variables and filtration defined in Appendix \ref{proof_of_mainlemma} and $c_j^{(s)}$ are the constant series defined in Definition \ref{defck}.

%We have the following proofs:
\subsection{Proof of Proposition \ref{gooddef}}
\begin{proof}[Proof of Proposition \ref{gooddef}]
 By the definition of reference point $\xb_t^{(s)}$ in \eqref{referencepoints}, we can easily verify that \eqref{gooddef_1} holds trivially.

 Next we prove \eqref{gooddef_2}. Note that by \eqref{gooddef_1} we have $\xb_{t}^{(s)} = \xb_{t'}^{(s)}$. For any $0 \leq s'\leq s$, it is also true that $\xb_{t}^{(s')} = \xb_{t'}^{(s')}$ by \eqref{referencepoints}, which means $\xb_{t}$ and $\xb_{t'}$ share the same first $s+1$ reference points. Then by the update rule of $\gb_{t}^{(s')}$ in Algorithm \ref{algorithm:1}, we will maintain $\gb_{t}^{(s')}$ unchanged from time step $t$ to ${t'}$. In other worlds, we have $\gb_{t}^{(s')} = \gb_{t'}^{(s')}$ for all $0 \leq s'\leq s$.

 We now prove the last claim \eqref{gooddef_3}. Based on \eqref{deftunv} and \eqref{gooddef_2}, we have $\vb_{t}^{(s)} = \sum_{s'=0}^s \gb_{t}^{(s')} = \sum_{s'=0}^s \gb_{p\cdot \prod_{j=s+1}^K T_j}^{(s')} = \vb_{p\cdot \prod_{j=s+1}^K T_j}^{(s)}$. Since for any $s \leq s'' \leq K$, we have the following equations by the update in Algorithm \ref{algorithm:1} (Line~\ref{alg_line:update_x}).
 \begin{align}
     \xb_{p\cdot \prod_{j=s+1}^K T_j}^{(s'')} &= \xb_{\lfloor p\cdot \prod_{j=s+1}^K T_j /\prod_{j=s''+1}^K T_j \rfloor \cdot \prod_{j=s''+1}^K T_j} \notag \\
     & =\xb_{ p\cdot \prod_{j=s+1}^K T_j /\prod_{j=s''+1}^K T_j  \cdot \prod_{j=s''+1}^K T_j} \notag \\
     & = \xb_{p\cdot \prod_{j=s+1}^K T_j}^{(s)}.\notag
 \end{align}
 Then for any $s<s'' \leq K$, we have
 \begin{align}
     \gb_{p\cdot \prod_{j=s+1}^K T_j}^{(s'')} = \frac{1}{B_{s''}}\sum_{i \in I}\bigg[\df_i\Big(\xb_{p\cdot \prod_{j=s+1}^K T_j}^{(s'')}\Big) - \df_i\Big(\xb_{p\cdot \prod_{j=s+1}^K T_j}^{(s''-1)}\Big)\bigg] = 0.\label{gooddef_4}
 \end{align}
Thus,  we have
\begin{align}
    \vb_{p\cdot \prod_{j=s+1}^K T_j} &= \sum_{s'' = 0}^K \gb_{p\cdot \prod_{j=s+1}^K T_j}^{(s'')}  = \sum_{s'' = 0}^s \gb_{p\cdot \prod_{j=s+1}^K T_j}^{(s'')} = \sum_{s'' = 0}^s \gb_{t}^{(s'')} = \vb_t^{(s)},\label{gooddef_5}
\end{align}
where the first equality holds because of the definition of $\vb_{p\cdot \prod_{j=s+1}^K T_j}$ , the second equality holds due to \eqref{gooddef_4} , the third equality holds due to \eqref{gooddef_2} and the last equality holds due to \eqref{deftunv}. This completes the proof of \eqref{gooddef_3}.
\end{proof}

\subsection{Proof of Lemma \ref{propck}}

\begin{proof}[Proof of Lemma \ref{propck}]
For any fixed $s$, it can be seen that from the definition in \eqref{defck_2}, $c_j^{(s)}$ is monotonically decreasing with $j$. In order to prove \eqref{propck_1}, we only need to compare $(1+T_{s-1})\cdot c_0^{(s-1)}$ and $ c^{(s)}_{T_s}$. Furthermore, by the definition of series $\{c^{(s)}_{j}\}$ in \eqref{defck_2}, it can be inducted that when $0 \leq j \leq T_{s-1}$,
\begin{align}
    c_j^{(s-1)} =
    \bigg(1+\frac{1}{T_{s-1}}\bigg)^{T_{s-1}-j}\cdot c_{T_{s-1}}^{(s-1)}+\frac{(1+1/T_{s-1})^{T_{s-1}-j}-1}{1/T_{s-1}}\cdot\frac{3L^2}{M}\cdot\frac{\prod_{l=s}^K T_l}{B_{s-1}}.\label{propck_7}
\end{align}
We take $j = 0$ in \eqref{propck_7} and obtain
\begin{align}
     c_0^{(s-1)}& = \bigg(1+\frac{1}{T_{s-1}}\bigg)^{T_{s-1}}\cdot c_{T_{s-1}}^{(s-1)}+\frac{(1+1/T_{s-1})^{T_{s-1}}-1}{1/T_{s-1}}\cdot\frac{3L^2}{M}\cdot\frac{\prod_{l=s}^K T_l}{B_{s-1}}\notag\\
     &<
     2.8\times c_{T_{s-1}}^{(s-1)}+\frac{6L^2}{M}\cdot\frac{\prod_{l=s-1}^K T_l}{B_{s-1}}\label{propck_3}\\
     &\leq
     \frac{2.8M+6L^2/M}{6^{K-s+2}\cdot \prod_{l=s-1}^K T_l}\label{propck_4}\\
     &<
     \frac{3M}{6^{K-s+2}\cdot \prod_{l=s-1}^K T_l},\label{propck_5}
\end{align}
where \eqref{propck_3} holds because $(1+1/n)^{n}<2.8$ for any $n \geq 1$, \eqref{propck_4} holds due to the definition of $c_{T_{s-1}}^{(s-1)}$ in \eqref{defck_1} and $B_{s-1} \geq 6^{K-s+2}(\prod_{l=s-1}^K T_l)^2$ and \eqref{propck_5} holds because $M\geq 6L$. Recall that $c_j^{(s)}$ is monotonically decreasing with $j$ and the inequality in \eqref{propck_5}. Thus for all $2\leq s\leq K$ and $0\leq j\leq T_{s-1}$, we have
\begin{align}
    (1+T_{s-1})\cdot c_j^{(s-1)} & \leq (1+T_{s-1})\cdot c_0^{(s-1)} \notag \\
    &\leq
    (1+T_{s-1})\cdot \frac{3M}{6^{K-s+2}\cdot \prod_{l=s-1}^K T_l}\notag \\
    & <
     \frac{6M}{6^{K-s+2}\cdot \prod_{l=s}^K T_l}\notag \\
     &= c^{(s)}_{T_s},\label{propck_6}
\end{align}
where the third inequality holds because $(1+T_{s-1})/T_{s-1}\leq 2$ when $T_{s-1}\geq 1$ and the last equation comes from the definition of $c_{T_s}^{s}$ in \eqref{defck_1}. This completes the proof of \eqref{propck_1}.

Using similar techniques, we can obtain the upper bound for $c_0^{K}$ which is similar to inequality \eqref{propck_5} with $s-1$ replaced by $K$. Therefore, we have
\begin{align*}
 (1+T_{K})\cdot c_j^{(K)} & \leq (1+T_{K})\cdot c_0^{(K)}   <
     \frac{6M}{6^{K-K+1}\cdot \prod_{l=K}^K T_l} \leq M,
\end{align*}
which completes the proof of \eqref{propck_2}.
\end{proof}

\subsection{Proof of Lemma \ref{oneloop}}
Now we prove Lemma \ref{oneloop}, which is a special case of Lemma \ref{muloop} if we choose $s = K$. 
\begin{proof}[Proof of Lemma \ref{oneloop}]
To simplify notations, we use $\EE[\cdot] $ to denote the conditional expectation $ \EE[\cdot|\cF_{p\cdot T_K}]$ in the rest of this proof. For $0 \leq j <T_K$, we denote $\hc = -(10M)^{-1}\cdot \vold$. According to the update in Algorithm \ref{algorithm:1} (Line~\ref{alg_line:update_gd}), we have
\begin{align}\label{eq:update_rule}
    \xnew= \xold+\hc,
\end{align}
which immediately implies
\begin{align}
    F(\xnew)
    &= F(\xold+\hc) \notag\\
    & \leq F(\xold)+ \la \dF(\xold), \hc\ra + \frac{L}{2}\|\hc\|_2^2\label{oneloop_1}\\
    & = \big[\la \vold, \hc\ra + 5M\|\hc\|_2^2\big] + F(\xold)\notag \\
    &\quad\quad + \la \dF(\xold) - \vold, \hc\ra + \bigg(\frac{L}{2}-5M\bigg)\|\hc\|_2^2\notag\\
    & \leq F(\xold)+ \la \dF(\xold) - \vold, \hc\ra + (L-5M)\|\hc\|_2^2,\label{oneloop_2}
\end{align}
where \eqref{oneloop_1} is due to the $L$-smoothness of $F$, which can be verified as follows 
\begin{align}
    \|\nabla F(\xb) - \nabla F(\yb)\|_2 &= \|\EE_i [\nabla f_i(\xb) - \nabla f_i(\yb)]\|_2\notag \\
    &\leq \sqrt{\EE_i\|\nabla f_i(\xb) - \nabla f_i(\yb)\|_2^2}\notag \\
    & \leq L\|\xb - \yb\|_2.\notag
\end{align}
%which is equivalent to Definition \ref{smooth}.
\eqref{oneloop_2} holds because $\la \vold, \hc\ra + 5M\|\hc\|_2^2  = -5M\|\hc\|_2^2\leq 0$. Further by Young's inequality, we obtain
\begin{align}
    F(\xnew)& \leq F(\xold) + \frac{1}{2M}\|\dF(\xold) - \vold\|_2^2 +\bigg(\frac{M}{2}+L-5M\bigg)\|\hc\|_2^2\notag\\
    & \leq F(\xold) + \frac{1}{M}\|\dF(\xold) - \vold\|_2^2  - 3M\|\hc\|_2^2,\label{oneloop_4}
\end{align}
where the second inequality holds because $M>L$. Now we bound the term $\cnew \|\xnew - \Kref\|_2^2$. By \eqref{eq:update_rule} we have
\begin{align*}
    \cnew\|\xnew - \Kref\|_2^2
    &= \cnew\|\xold - \Kref+\hc\|_2^2 \\
    & = \cnew\big[\|\xold - \Kref\|_2^2+\|\hc\|_2^2+2\la \xold - \Kref, \hc \ra\big]. 
\end{align*}
Applying Young's inequality yields
\begin{align}
    \cnew\|\xnew - \Kref\|_2^2 & \leq \cnew\bigg[\|\xold - \Kref\|_2^2+\|\hc\|_2^2\notag\\
    &\qquad +\frac{1}{T_K}\| \xold - \Kref\|_2^2 + T_K\|\hc \|_2^2\bigg] \notag\\
    & = \cnew \bigg[\bigg(1+\frac{1}{T_K}\bigg)\|\xold - \Kref\|_2^2+(1+T_K)\|\hc\|_2^2\bigg],\label{oneloop_6}
\end{align}
Adding up inequalities \eqref{oneloop_6} and \eqref{oneloop_4}, we get
\begin{align}
    &F(\xnew) + \cnew\|\xnew - \Kref\|_2^2 \notag\\
    &\leq F(\xold) + \frac{1}{M}\|\dF(\xold) - \vold\|_2^2 - \big[3M - \cnew(1+T_K)\big]\|\hc\|_2^2\notag \\
   & \quad\quad +\cnew \bigg(1+\frac{1}{T_K}\bigg)\|\xold - \Kref\|_2^2\notag\\
    & \leq
    F(\xold) + \frac{1}{M}\|\dF(\xold) - \vold\|_2^2 - 2M\|\hc\|_2^2\notag\\
    &\qquad+\cnew \bigg(1+\frac{1}{T_K}\bigg)\|\xold - \Kref\|_2^2\label{oneloop_7},
\end{align}
where the last inequality holds due to the fact that $\cnew(1+T_K)<M$ by Lemma \ref{propck}.
Next we bound $\|\dF(\xold)\|_2^2$ with $\|\hc\|_2^2$.  Note that by \eqref{eq:update_rule}, we have
\begin{align}
    \|\dF(\xold)\|_2^2 %&= \big\|\dF(\xold) - \big[\vold+10M\hc\big]\big\|_2^2\notag\\
    & = \big\|\big[\dF(\xold) - \vold\big] - 10M\hc\big\|_2^2\notag\\
    & \leq 2\big(\|\dF(\xold) - \vold\|_2^2+100M^2\|\hc\|_2^2\big),\notag
\end{align}
which immediately implies
\begin{align}
    -2M\|\hc\|_2^2\leq\frac{2}{100M}\big(\|\dF(\xold) -  \vold\|_2^2-\frac{1}{100M}\|\dF(\xold)\|_2^2.\label{oneloop_8}
\end{align}
Plugging \eqref{oneloop_8} into \eqref{oneloop_7}, we have
\begin{align}
    &F(\xnew) + \cnew\|\xnew - \Kref\|_2^2 \notag\\
    & \leq
    F(\xold) + \frac{1}{M}\|\dF(\xold) - \vold\|_2^2  + \frac{1}{50M}\cdot \|\dF(\xold) - \vold\|_2^2 \notag\\
    &\quad\quad- \frac{1}{100M}\|\dF(\xold)\|_2^2 +\cnew \bigg(1+\frac{1}{T_K}\bigg)\|\xold - \Kref\|_2^2\notag\\
    & \leq
    F(\xold) + \frac{2}{M} \|\dF(\xold) - \vold\|_2^2 - \frac{1}{100M}\|\dF(\xold)\|_2^2  \notag\\
    &\qquad+\cnew \bigg(1+\frac{1}{T_K}\bigg)\|\xold - \Kref\|_2^2.\label{oneloop_10}
\end{align}
Next we bound $\|\dF(\xold) - \vold\|_2^2$. First, by Lemma \ref{inductgra} we have
\begin{align}
    \ES\Big\|\dF(\xb_{p\cdot T_K+j}^{(K)}) - \vb_{p\cdot T_K+j}^{(K)}\Big\|_2^2 \leq \frac{L^2}{B_K}  \EE\Big\|\xb_{p\cdot T_K+j}^{(K)} - \xb_{p\cdot T_K+j}^{(K-1)}\Big\|_2^2 + \EE\Big\|\dF(\xb_{p\cdot T_K+j}^{(K-1)}) - \vb_{p\cdot T_K+j}^{(K-1)}\Big\|_2^2.\notag
\end{align}
Since $ \xb_{p\cdot T_K+j}^{(K)} = \xold,  \vb_{p\cdot T_K+j}^{(K)} = \vold$, $\xb_{p\cdot T_K+j}^{(K-1)} = \xb_{p\cdot T_K}$ and $\vb_{p\cdot T_K+j}^{(K-1)} = \vb_{p\cdot T_K}$, we have
\begin{align}
\ES\|\dF(\xold) - \vold\|_2^2 \leq \frac{L^2}{B_K}  \EE\|\xold - \Kref\|_2^2 + \EE\|\dF(\Kref) - \vkm\|_2^2. \label{oneloop_9}
\end{align}
We now
take expectation $\EE[\cdot]$ with \eqref{oneloop_10} and plug \eqref{oneloop_9} into \eqref{oneloop_10}.  We obtain that 
\begin{align}
    &\ES\bigg[F(\xnew) + \cnew\|\xnew - \Kref\|_2^2 +\frac{1}{100M}\|\dF(\xold)\|_2^2\bigg]\notag\\
    & \leq
    \EE\bigg[F(\xold) + \bigg(\cnew\bigg(1+\frac{1}{T_K}\bigg)+ \frac{3L^2}{B_KM}\bigg)\|\xold - \Kref\|_2^2 +\frac{2}{M} \|\dF(\Kref) - \vkm\|_2^2\bigg]\notag\\
    & =
    \EE\bigg[F(\xold) + \cold\|\xold - \Kref\|_2^2+\frac{2}{M}\cdot \|\dF(\Kref) - \vkm\|_2^2\bigg],\label{oneloop_11}
\end{align}
where \eqref{oneloop_11} holds because we have $\cold = \cnew(1+1/T_K)+ 3L^2/(B_K M)$ by Definition \ref{defck}. Telescoping \eqref{oneloop_11} for $j = 0$ to $T_K-1$, we have
\begin{align}
    &\ES \big[F\big(\xb_{(p+1)\cdot T_K}\big)+c_{T_K}^{(K)}\cdot \|\xb_{(p+1)\cdot T_K} - \xb_{p\cdot T_K}\|_2^2\big]+\frac{1}{100M}\sum_{j=0}^{T_{K}-1}\ES \|\dF(\xb_{p\cdot T_K+j})\|_2^2\notag\\
    &\leq F(\xb_{p\cdot T_K})+\frac{2T_K}{M}\cdot\EE \|\dF(\xb_{p\cdot T_K}) - \vb_{p\cdot T_K}\|_2^2,
\end{align}
which completes the proof.

\end{proof}

\iffalse
\begin{proof}[Another proof]
$\hc = \vold/M$, then
\begin{align}
    \EE F(\xnew) &\leq \EE\big[F(\xold)-1/M\cdot \la \vold , \dF(\xold)\ra + L/2\cdot 1/M^2\cdot \|\vold\|_2^2\notag \\
    & =
    \EE F(\xold) - 1/M\cdot \EE\|\dF(\xold)\|_2^2+L/2\cdot 1/M^2\cdot\EE\|\vold\|_2^2.
\end{align}
We have
\begin{align}
    &\cnew\|\xnew - \Kref\|_2^2 \notag \\
    &= \cnew\|\xold - \Kref+\hc\|_2^2\notag \\
    & = \cnew\big[\|\xold - \Kref\|_2^2+\|\hc\|_2^2+2\la \xold - \Kref, \hc \ra\big] \notag\\
    & \leq \cnew\big[\|\xold - \Kref\|_2^2+\|\hc\|_2^2+\| \xold - \Kref\|_2^2 / T_K+  T_K\|\hc \|_2^2\big] \\
    & = \cnew \big[(1+1/T_K)\|\xold - \Kref\|_2^2+(1+T_K)\|\hc\|_2^2\big],\\
    & = \cnew \big[(1+1/T_K)\|\xold - \Kref\|_2^2+(1+T_K)/M^2\cdot\|\vold\|_2^2\big],
\end{align}
We have
\begin{align}
    &\EE\big[F(\xnew)+\cnew\|\xnew - \Kref\|_2^2\big]\notag \\
    & \leq
    \EE\big[F(\xold) - 1/M\cdot \|\dF(\xold)\|_2^2+1/M^2\cdot(L/2+\cnew(1+T_K))\cdot \|\vold\|_2^2\notag \\
    &\quad\quad +\cnew(1+1/T_K)\|\xold - \Kref\|_2^2\big]\notag \\
    & \leq
    \EE\big[F(\xold) - 1/M\cdot \|\dF(\xold)\|_2^2+\cnew(1+1/T_K)\|\xold - \Kref\|_2^2\big]\notag \\
    &\quad\quad +3L/M^2\cdot \EE\|\vold\|_2^2\notag \\
    & =
   \EE\big[F(\xold) - 1/M\cdot \|\dF(\xold)\|_2^2+\cnew(1+1/T_K)\|\xold - \Kref\|_2^2\big]\notag \\
    &\quad\quad +3L/M^2\cdot\big[ \EE\|\vold - \dF(\xold)\|_2^2 + \|\dF(\xold)\|_2^2\big]\notag\notag
\end{align}

\end{proof}
\fi
%inductive loop

\subsection{Proof of Lemma \ref{inductgra}}

\begin{proof}[Proof of Lemma \ref{inductgra}]
%For simplification, we denote $\EE[\cdot] =\EE[\cdot|\cF_{t^l}]$ in the rest of this proof.
If $t^l = t^{l-1}$, we have $\xb_t^{(l)} = \xb_t^{(l-1)}$ and $\vb_t^{(l)} = \vb_t^{(l-1)}$. In this case the statement in Lemma \ref{inductgra} holds trivially. Therefore, we assume $t^{l} \neq t^{l-1}$ in the following proof. Note that
\begin{align}
    &\Ep \big[\big\|\dF(\xb_t^{(l)}) - \vb_t^{(l)}\big\|_2^2|\cF_{t^l} \big]\notag\\
    &= \Ep \big[\big\|\dF(\xb_t^{(l)}) - \vb_t^{(l)} - \Ep\big[\dF(\xb_t^{(l)}) - \vb_t^{(l)}\big]\big\|_2^2|\cF_{t^l}\big] + \big\|\Ep\big[\dF(\xb_t^{(l)}) - \vb_t^{(l)}|\cF_{t^l}\big]\big\|_2^2 \notag\\
    & =
    \underbrace{\Ep\Bigg[ \bigg\|\dF(\xb_t^{(l)}) - \sum_{j=0}^l \gb_t^{(j)} - \Ep\bigg[\dF(\xb_t^{(l)}) - \sum_{j=0}^l \gb_t^{(j)}\bigg]\bigg\|_2^2\bigg|\cF_{t^l}\bigg]}_{J_1} +
    \underbrace{\bigg\|\EE\bigg[\dF(\xb_t^{(l)})-\sum_{j=0}^{l} \gb_t^{(j)}\bigg|\cF_{t^l}\bigg]  \bigg\|_2^2}_{J_2},\label{inductgra_0.9}
\end{align}
where in the second equation we used the definition $\vb_t^{(l)} = \sum_{i=0}^l \gb_t^{(i)}$  in \eqref{deftunv}. We first upper bound term $J_1$. According to the update rule in Algorithm \ref{algorithm:1} (Line~\ref{alg_line:update_grad_small}-\ref{alg_line:update_grad_large}), when $j<l$, $\gb_t^{(j)}$ will not be updated at the $t^l$-th iteration. Thus we have $\EE [\gb_t^{(j)}|\cF_{t^l}] = \gb_t^{(j)}$ for all $j<l$. In addition, by the definition of $\cF_{t^l}$, we have $\EE [\dF(\xb_t^{(l)})|\cF_{t^l}] = \dF(\xb_t^{(l)}) $. Then we have the following equation
\begin{align}\label{eq:J1_gt_Egt}
  J_1=\Ep\big[\big\|\gb_t^{(l)} - \Ep \big[\gb_t^{(l)}|\cF_{t^l}\big] \big\|_2^2|\cF_{t^l}\big].
\end{align}
We further have
\begin{align}
\gb_t^{(l)} =\frac{1}{B_l}\sum_{i \in I}\big[\df_i(\xb_t^{(l)}) - \df_i(\xb_t^{(l-1)})\big],\quad \EE\big[\gb_t^{(l)}\big|\cF_{t^l}\big] =\dF(\xb_t^{(l)}) - \dF(\xb_t^{(l-1)}).
\end{align}
Therefore, we can apply Lemma \ref{concen} to \eqref{eq:J1_gt_Egt} and obtain
\begin{align}
 J_1&\leq \frac{1}{B_l}\cdot \frac{1}{n}\sum_{i =1}^n\big\|\df_i(\xb_t^{(l)}) - \df_i(\xb_t^{(l-1)})-\big[\dF(\xb_t^{(l)}) - \dF(\xb_t^{(l-1)})\big]\big\|_2^2\notag  \\
 &\leq\frac{1}{B_l n}\sum_{i =1}^n \big\|\df_i(\xb_t^{(l)}) - \df_i(\xb_t^{(l-1)})\big\|_2^2\notag\\
 & \leq\frac{L^2}{B_l}\big\|\xb_t^{(l)} - \xb_t^{(l-1)}\big\|_2^2,\label{inductgra_4}
\end{align}
where the second inequality is due to the fact that $\EE[\|\bX-\EE[\bX]\|_2^2]\leq\EE\|\bX\|_2^2$ for any random vector $\bX$ and the last inequality holds due to the fact that $F$ has averaged $L$-Lipschitz gradient.

Next we turn to bound term $J_2$. Note that
\begin{align}
    \Ep \big[\gb_t^{(l)}\big|\cF_{t^l}\big]  = \EE \bigg[\frac{1}{B_l}\sum_{i \in I}\big[\df_i(\xb_t^{(l)}) - \df_i(\xb_t^{(l-1)})\big]\bigg|\cF_{t^l}\bigg] =  \dF(\xb_t^{(l)}) - \dF(\xb_t^{(l-1)}),\notag
\end{align}
which immediately implies
\begin{align}
     \EE\bigg[\dF(\xb_t^{(l)})-\sum_{j=0}^{l} \gb_t^{(j)}\bigg|\cF_{t^l}\bigg]&=\EE\bigg[\dF(\xb_t^{(l)}) - \dF(\xb_t^{(l)}) + \dF(\xb_t^{(l-1)}) - \sum_{j=0}^{l-1}   \gb_t^{(j)}\bigg|\cF_{t^l}\bigg] \notag\\
     &= \EE\big[\dF(\xb_t^{(l-1)}) - \vb_t^{(l-1)}\big|\cF_{t^l}\big]\\
     &=\dF(\xb_t^{(l-1)}) - \vb_t^{(l-1)},\notag
\end{align}
where the last equation is due to the definition of $\cF_t$. Plugging $J_1$ and $J_2$ into \eqref{inductgra_0.9} yields the following result:
\begin{align}
    \EE\big[ \big\|\dF(\xb_t^{(l)}) - \vb_t^{(l)}\big\|_2^2 \big|\cF_{t^l}\big]\leq \frac{L^2}{B_l}\big\|\xb_t^{(l)} - \xb_t^{(l-1)}\big\|_2^2 + \big\|\dF(\xb_t^{(l-1)}) - \vb_t^{(l-1)}\big\|_2^2,
\end{align}
which completes the proof.
\end{proof}

\section{Additional Experimental Results}

\begin{figure}[t!]
%\vskip -0.2in
	\begin{center}
		\subfigure[training loss (\textit{MNIST})]{\includegraphics[width=0.32\linewidth]{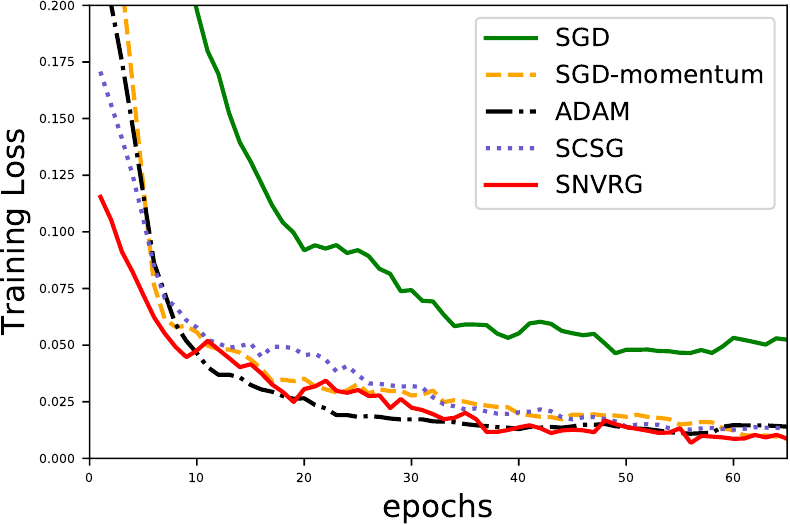}
		\label{no_mnist_loss}}		
		\subfigure[training loss (\textit{CIFAR10})]{\includegraphics[width=0.32\linewidth]{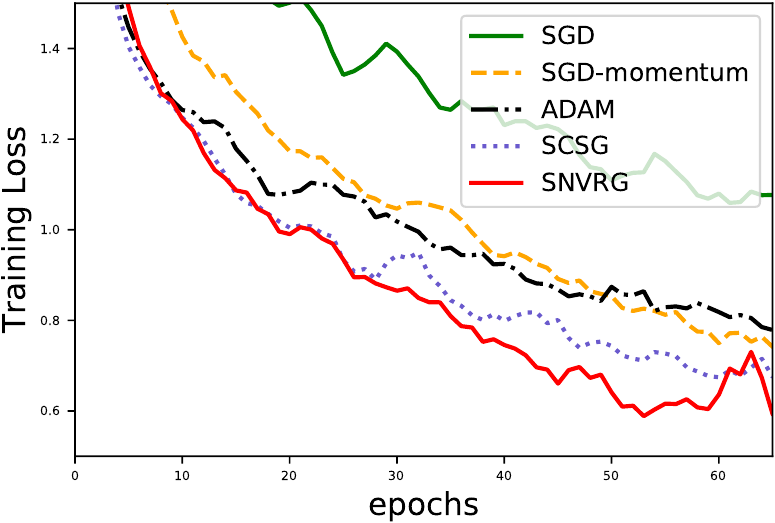}\label{no_cifar10_loss}
		}
    	\subfigure[training loss (\textit{SVHN})]{\includegraphics[width=0.32\linewidth]{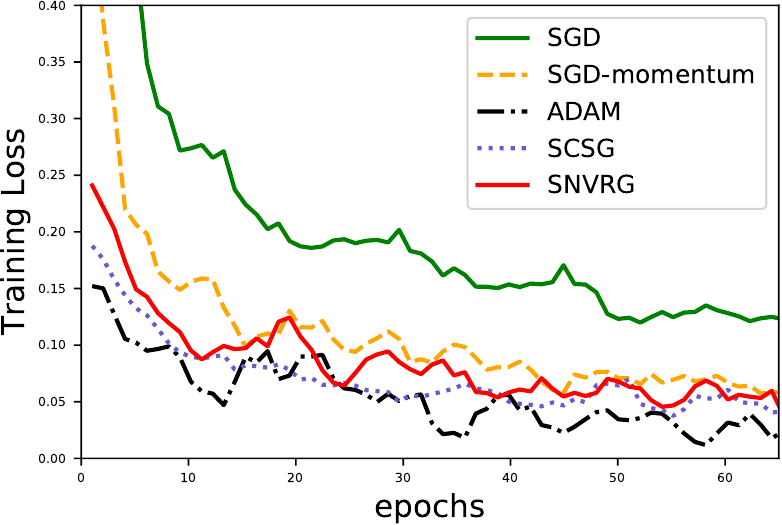}\label{no_svhn_loss}
    	}		
		
		\subfigure[test error (\textit{MNIST})]{\includegraphics[width=0.32\linewidth]{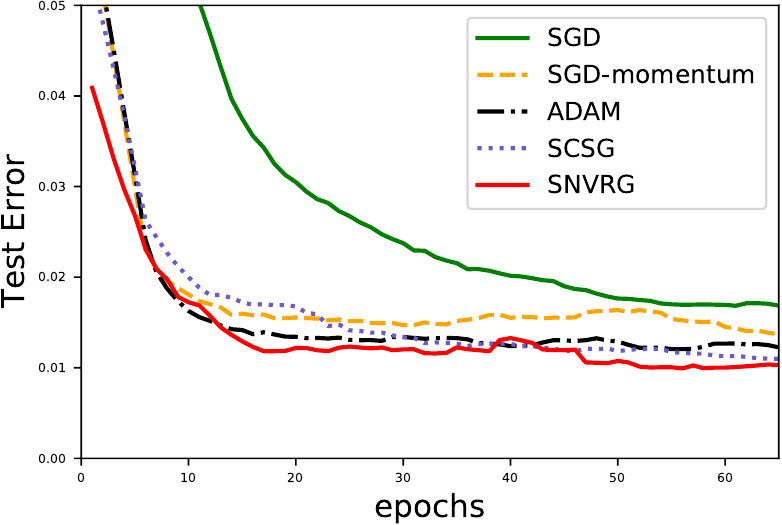}
		\label{no_mnist_test}}		
		\subfigure[test error (\textit{CIFAR10})]{\includegraphics[width=0.32\linewidth]{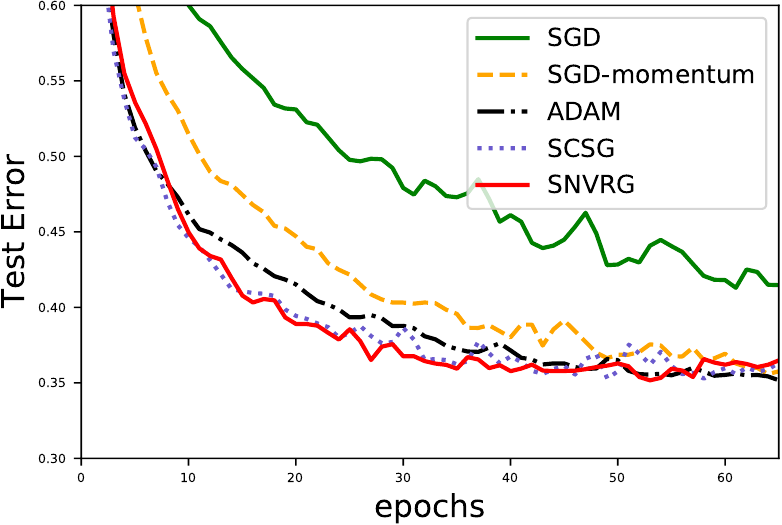}\label{no_cifar10_test}}
    	\subfigure[test error (\textit{SVHN})]{\includegraphics[width=0.32\linewidth]{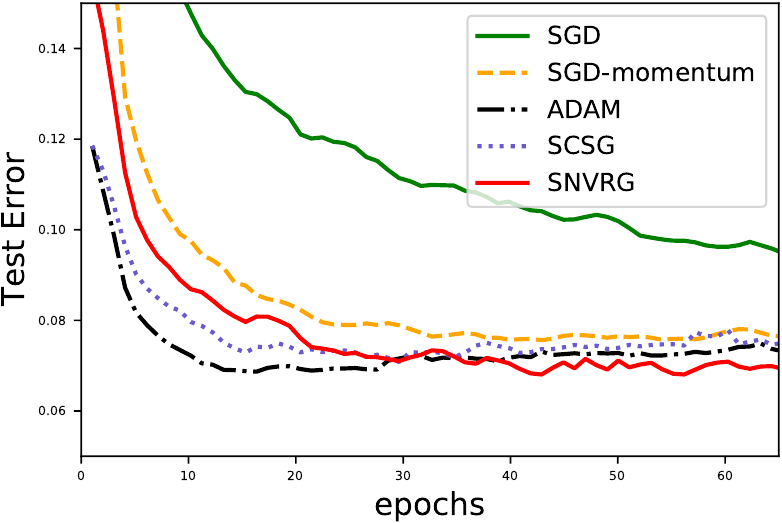}
    	\label{no_svhn_test}}
		
	\caption{Experimental results on different datasets without learning rate decay. (a) and (d) depict the training loss and test error (top-1 error) v.s. data epochs for training LeNet on MNIST dataset. (b) and (e) depict the training loss and test error v.s. data epochs for training LeNet on CIFAR10 dataset. (c) and (f) depict the training loss and test error v.s. data epochs for training LeNet on SVHN dataset. \label{no_fig_result}}
	\end{center}
	%\vskip -0.1in
\end{figure}

We also conducted experiments comparing different algorithms without the learning rate decay schedule. The parameters are tuned by the same grid search described in Section \ref{sec:experiment}. In particular, we summarize the parameters of different algorithms used in our experiments with and without learning rate decay for MNIST in Table \ref{table:mnist}, CIFAR10 in Table \ref{table:cifar10}, and SVHN in Table \ref{table:svhn}. 
We plotted the training loss and test error for each dataset without learning rate decay in Figure \ref{no_fig_result}. The results on MNIST are presented in Figures \ref{no_mnist_loss} and \ref{no_mnist_test}; the results on CIFAR10 are in Figures \ref{no_cifar10_loss} and \ref{no_cifar10_test}; and the results on SVHN dataset are shown in Figures \ref{no_svhn_loss} and \ref{no_svhn_test}. It can be seen that without learning decay, our algorithm \emph{SNVRG} still outperforms all the baseline algorithms except for the training loss on SVHN dataset. However, \emph{SNVRG} still performs the best in terms of test error on SVHN dataset. These results again suggest that \emph{SNVRG} can beat the state-of-the-art in practice, which backups our theory.

%\CC{Change No to N/A in the tables}

\begin{table}[h]
\caption{Parameter settings of all algorithms on MNIST dataset. %$\eta$ is the learning rate, $B$ is the batch size and $b$ is the batch size ratio. In the learning rate decay setting, the learning rate $\eta$ is the initial rate. 
}
\label{table:mnist}
\begin{small}
\begin{center}
\begin{tabular}{ccccccc}
\toprule
\multirow{3}{*}{Algorithm }& \multicolumn{3}{c}{With Learning Rate Decay}   & \multicolumn{3}{c}{Without Learning Rate Decay} \\
\cmidrule{2-7}
& Initial learning &Batch size  & Batch size  & learning &Batch size  & Batch size\\
 & rate $\eta$ &$B$ & ratio $b$ & rate $\eta$& $B$ &  ratio $b$\\
%& $\eta$ (initial) & $B$ &  $b$  &  $\eta$& $B$ &  $b$\\
%&&&\\
\midrule
SGD&0.1 & 1024 & N/A &0.01 & 1024 & N/A\\
SGD-momentum&0.01 & 1024 & N/A &0.1 & 1024 & N/A\\
ADAM&0.001 & 1024 & N/A &0.001 & 1024 & N/A\\
SCSG &0.01 & 512 & 8 &0.01 & 512 & 8\\
SNVRG  &0.01 & 512 & 8 &0.01 & 512 & 8\\
\bottomrule
\end{tabular}
\end{center}
\end{small}
\end{table}

\begin{table}[h]
\caption{Parameter settings of all algorithms on CIFAR10 dataset.}
\label{table:cifar10}
\begin{small}
\begin{center}
\begin{tabular}{ccccccc}
\toprule
\multirow{3}{*}{Algorithm }& \multicolumn{3}{c}{With Learning Rate Decay}   & \multicolumn{3}{c}{Without Learning Rate Decay} \\
\cmidrule{2-7}
& Initial learning &Batch size  & Batch size  & learning &Batch size  & Batch size\\
 & rate $\eta$ &$B$ & ratio $b$ & rate $\eta$& $B$ &  ratio $b$\\
%&&&\\
\midrule
SGD&0.1 & 1024 & N/A &0.01 & 512 & N/A\\
SGD-momentum&0.01 & 1024 & N/A &0.01 & 2048 & N/A\\
ADAM&0.001 & 1024 & N/A &0.001 & 2048 & N/A\\
SCSG &0.01 & 512 & 8 &0.01 & 512 & 8\\
SNVRG  &0.01 & 1024 & 8 &0.01 & 512 & 4\\
\bottomrule
\end{tabular}
\end{center}
\end{small}
\end{table}

\begin{table}[h]
\caption{Parameter settings of all algorithms on SVHN dataset.}
\label{table:svhn}
\begin{small}
\begin{center}
\begin{tabular}{ccccccc}
\toprule
\multirow{3}{*}{Algorithm }& \multicolumn{3}{c}{With Learning Rate Decay}   & \multicolumn{3}{c}{Without Learning Rate Decay} \\
\cmidrule{2-7}
& Initial learning &Batch size  & Batch size  & learning &Batch size  & Batch size\\
 & rate $\eta$ &$B$ & ratio $b$ & rate $\eta$& $B$ &  ratio $b$\\
%&&&\\
\midrule
SGD&0.1 & 2048 & N/A &0.01 & 1024 & N/A\\
SGD-momentum&0.01 & 2048 & N/A &0.01 & 2048 & N/A\\
ADAM&0.001 & 1024 & N/A &0.001 & 512 & N/A\\
SCSG &0.01 & 512 & 4 &0.1 & 1024 & 4\\
SNVRG  &0.01 & 512 & 8 &0.01 & 512 & 4\\
\bottomrule
\end{tabular}
\end{center}
\end{small}
\end{table}

\section{An Equivalent Version of Algorithm \ref{algorithm:1}}
Recall the One-epoch-$\algname$ algorithm in Algorithm \ref{algorithm:1}. Here we present an equivalent version of Algorithm \ref{algorithm:1} using nested loops, which is displayed in Algorithm \ref{algorithm:equivalent} and is more aligned with the illustration in Figure \ref{fig:illu_snvrg}. Note that the notation used in Algorithm \ref{algorithm:equivalent} is slightly different from that in Algorithm \ref{algorithm:1} to avoid confusion. %We would like to emphasize that Algorithms \ref{algorithm:1} and \ref{algorithm:equivalent} are essentially the same algorithm. %Now, we are going to reorganize Algorithm \ref{algorithm:1} into a nested $K$-loop algorithm more rigorously.

\begin{algorithm*}[t!]
\caption{One-epoch SNVRG($F, \xb_0, K, M, \{T_i\}, \{B_i\}, B$)}\label{algorithm:equivalent}
\begin{algorithmic}[1]
   \STATE \textbf{Input:} Function $F$, starting point $\xb_0$, loop number $K$, step size parameter $M$,  loop parameters $T_i, i \in [K]$, batch parameters $B_i, i \in [K]$, base batch $B>0$. \\
   \textbf{Output:} $[\xb_{\text{out}}, \xb_{\text{end}}]$

  \STATE $T\leftarrow \prod_{l=1}^K T_l$
  \STATE Uniformly generate index set $I \subset [n]$ without replacement
  \STATE $\gb_{[t_0]}^{(0)} \leftarrow \frac{1}{B}\sum_{i \in I}\nabla f_{i_d}(\xb_0)$
  \STATE  $\xb_{[0]}^{(l)} \gets \xb_0,\quad 0 \leq l \leq K$,

  \FOR{$t_1 = 0,\dots,T_1-1$}
  \STATE Uniformly generate index set $I \subset [n]$ without replacement, $|I| = B_1$
  \STATE $\gb_{[t_1]}^{(1)} \gets \frac{1}{B_1}\sum_{i \in I} \big[\nabla f_{i}(\xb_{[t_1]}^{(1)}) - \nabla f_{i}(\xb_{[0]}^{(0)})\big]$

  \STATE $\dots$
  \FOR{$t_l = 0,\dots,T_l-1$}
  \STATE Uniformly generate index set $I \subset [n]$ without replacement, $|I| = B_l$
  \STATE $\gb_{[t_l]}^{(l)} \gets \frac{1}{B_l}\sum_{i \in I} \big[\nabla f_{i}(\xb_{[t_l]}^{(l)}) - \nabla f_{i}(\xb_{[t_{l-1}]}^{(l-1)})\big]$

  \STATE $\dots$
  \FOR{$t_K = 0,\dots,T_K-1$}
  \STATE Uniformly generate index set $I \subset [n]$ without replacement, $|I| = B_K$
  \STATE $\gb_{[t_K]}^{(K)} \gets \frac{1}{B_K}\sum_{i \in I} \big[\nabla f_{i}(\xb_{[t_K]}^{(K)}) - \nabla f_{i}(\xb_{[t_{K-1}]}^{(K-1)})\big]$
  \STATE Denote $t = \sum_{j=1}^K t_j \prod_{l=j+1}^K T_l$, then let $\xb_{t+1} \gets \xb_t- 1/(10M)\cdot \sum_{l=0}^K \gb_{[t_l]}^{(l)}$
  \STATE $\xb_{[t_K+1]}^{(K)} \gets \xb_{t+1}$
  \ENDFOR

  \STATE $\dots$
  \STATE $\xb_{[t_l+1]}^{(l)} \gets \xb_{[T_{l+1}]}^{(l+1)}$
  \ENDFOR

  \STATE $\dots$

  \STATE $\xb_{[t_1+1]}^{(1)} \gets \xb_{[T_2]}^{(2)}$
  \ENDFOR
  \STATE  $\xb_{\text{out}}\leftarrow$ a uniformly random choice from $\{\xb_0, ..., \xb_{T-1}\}$
  %\STATE  $\xb_{\text{end}}\leftarrow\xb_T$
  \RETURN $[\xb_{\text{out}}, \xb_{T}]$

\end{algorithmic}
\end{algorithm*}

\end{document}